\documentclass[preprint,authoryear,10pt,a4paper]{elsarticle}
\usepackage[top=0.75in, bottom=0.75in, left=0.75in, right=0.75in]{geometry}
\usepackage{palatino}
\usepackage{amsmath,amsthm}
\usepackage{amssymb}
\usepackage{bbm}
\usepackage{amsfonts}
\usepackage{graphicx}
\usepackage{caption}
\usepackage{subcaption}
\usepackage{grffile}
\graphicspath{{graphics/}}
\usepackage[ruled,vlined,linesnumbered]{algorithm2e}
\usepackage{tabularx,booktabs}
\usepackage{inputenc} 
\usepackage[T1]{fontenc}   
\usepackage{url}            
\usepackage{booktabs}       %
\usepackage{amsmath,amsthm}
\usepackage{amssymb}
\usepackage{bbm}
\usepackage{amsfonts}       
\usepackage{nicefrac}       
\usepackage{microtype}      
\usepackage{tikz}
\usepackage{rotfloat}
\usepackage{graphicx}
\usepackage{caption}
\usepackage[ruled,vlined]{algorithm2e}
\usepackage{algpseudocode}
\usepackage{setspace}
\usepackage{appendix}
\usepackage{tablefootnote}
\usepackage{bm}
\usepackage{mathrsfs}
\usepackage{xcolor}
\usepackage{colortbl}

\usepackage{verbatim}
\usepackage{xcolor}
\definecolor{darkblue}{rgb}{0.0,0.5,0.5}
\definecolor{blue}{rgb}{0.0,0.59,0.84}
\definecolor{myblue}{RGB}{0,0,255}
  
\usepackage[colorlinks]{hyperref}
\hypersetup{colorlinks,breaklinks,linkcolor=blue,urlcolor=blue,anchorcolor=blue,citecolor=blue}
\usepackage{booktabs,caption}
\usepackage{multirow,bigdelim}
\usepackage{lscape}
\usepackage{lineno}
\usepackage{microtype}
\newtheorem*{remark}{Remark}

\usepackage{soul}
\usepackage{lineno}
\usepackage{fancyhdr}

\usepackage{amsmath,amsfonts,bm}

\def\eqref#1{equation~\ref{#1}}

\def\1{\bm{1}}

\def\ra{{\textnormal{a}}}

\def\vtheta{{\bm{\theta}}}

\def\vb{{\bm{b}}}

\def\ve{{\bm{e}}}

\def\vh{{\bm{h}}}

\def\vm{{\bm{m}}}

\def\vu{{\bm{u}}}
\def\vv{{\bm{v}}}

\def\vx{{\bm{x}}}

\def\mA{{\bm{A}}}

\def\mD{{\bm{D}}}
\def\mE{{\bm{E}}}

\def\mH{{\bm{H}}}
\def\mI{{\bm{I}}}

\def\mM{{\bm{M}}}

\def\mS{{\bm{S}}}

\def\mU{{\bm{U}}}
\def\mV{{\bm{V}}}
\def\mW{{\bm{W}}}
\def\mX{{\bm{X}}}

\DeclareMathAlphabet{\mathsfit}{\encodingdefault}{\sfdefault}{m}{sl}
\SetMathAlphabet{\mathsfit}{bold}{\encodingdefault}{\sfdefault}{bx}{n}

\def\gG{{\mathcal{G}}}

\newcommand{\E}{\mathbb{E}}

\newcommand{\normlp}{L^p}

\journal{Transportation Research Part E: Logistics and Transportation Review}

\begin{document}

\begin{frontmatter}





\title{{\fontfamily{lmss}\selectfont {Joint Estimation and Prediction of City-wide Delivery Demand: A Large Language Model Empowered Graph-based Learning Approach}}}



\author[label,label1]{Tong Nie}
\author[label]{Junlin He}
\author[label1]{Yuewen Mei}
\author[label1]{Guoyang Qin}
\author[label]{Guilong Li}
\author[label1]{Jian Sun\corref{cor1}}
\ead{sunjian@tongji.edu.cn}
\author[label]{Wei Ma\corref{cor1}}
\ead{wei.w.ma@polyu.edu.hk}

\address[label]{Department of Civil and Environmental Engineering, The Hong Kong Polytechnic University, Hong Kong SAR, China}
\address[label1]{Department of Traffic Engineering, Tongji University, Shanghai, 201804, China}

\cortext[cor1]{Corresponding authors.}

\begin{abstract}

The proliferation of e-commerce and urbanization has significantly intensified delivery operations in urban areas, boosting the volume and complexity of delivery demand.
Data-driven predictive methods, especially those utilizing machine learning techniques, have emerged to handle these complexities in urban delivery demand management problems. 
{One particularly pressing issue that has yet to be sufficiently addressed is the joint estimation and prediction of city-wide delivery demand, as well as the generalization of the model to new cities. }
To this end, we formulate this problem as a {transferable} graph-based spatiotemporal learning task. 
First, an {individual-collective} message-passing neural network model is formalized to capture the interaction between demand patterns of associated regions.
Second, by exploiting recent advances in large language models (LLMs), {we extract general geospatial knowledge encodings from the unstructured locational data using the embedding generated by LLMs.}
Last, to encourage the cross-city generalization of the model, {we integrate the encoding into the demand predictor in a transferable way.}
Comprehensive empirical evaluation results on two real-world delivery datasets, including eight cities in China and the US, demonstrate that our model significantly outperforms state-of-the-art baselines {in accuracy, efficiency, and transferability.}
\textbf{PyTorch implementation is available at:} \url{https://github.com/tongnie/IMPEL}.
\end{abstract}


\end{frontmatter}


\section{Introduction}
\begin{figure}[!htbp]
  \centering
  \captionsetup{skip=1pt}
  \includegraphics[width=1\columnwidth]{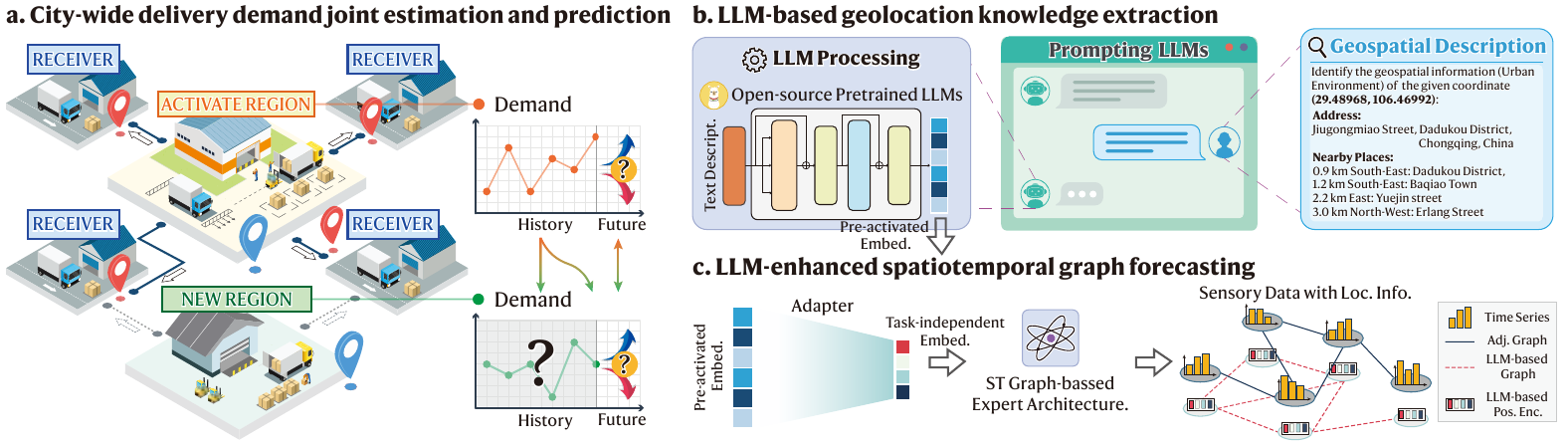}
  \caption{{This paper studies the city-wide delivery demand joint estimation and prediction problem, with the objective of estimating the demands for new regions and predicting the future demands of both existing and developing regions. We address three key challenges: (1) modeling the interaction between demand patterns of correlated regions, (2) integrating unstructured geospatial information into the demand estimator/predictor, and (3) transferring the model to new cites without re-training. The solution is developed on the foundation of an LLM-based geolocation knowledge extraction module and an LLM-enhanced spatiotemporal graph forecasting architecture.}}
  \label{fig:intro}
\end{figure}

Driven by the rapid expansion of e-commerce and the concomitant increase in urban freight traffic, urban delivery demand management is of paramount significance due to its profound implications for various aspects of urban life, including urban mobility, environmental sustainability, and economic efficiency, emerging as a critical area of inquiry in transportation research.
The increasing volume of deliveries in metropolitan areas has exacerbated traffic congestion, leading to longer travel times, higher fuel consumption, and greater greenhouse gas emissions \citep{yannis2006effects}. As e-commerce continues to proliferate, the volume and complexity of delivery operations in metropolitan areas have surged, necessitating more sophisticated and responsive management strategies. Thus, efficient management of delivery demand is crucial to alleviating these adverse effects, ensuring that delivery operations are both effective and sustainable \citep{black2010understanding,srinivas2022autonomous}.

To efficiently manage delivery demand in urban logistics systems, a variety of specific tasks are studied to improve operational efficiency and sustainability, such as parking time optimization \citep{reed2024does}, route planning \citep{he2022route}, system architecture design \citep{azcuy2021designing}, and strategic design \citep{janjevic2019integrating}. These tasks rely heavily on the accurate and timely estimation \citep{munuzuri2012estimation} and forecasting of the delivery demand \citep{hess2021real,nuzzolo2014urban}.
The significance of this issue is further underscored by several pressing challenges, including the high variability in demand patterns, the necessity for dynamic responsiveness in logistics operations, and precision in logistics planning. 
In this study, the term \textit{``delivery demand''} is defined as the total number of pickup (or dropoff) delivery orders, including package and food orders, within a specified region of interest (ROI). Subsequently, the spatio-temporal evolution and distribution of city-wide demand patterns of all ROIs are examined.

Traditional approaches to demand forecasting and estimation have often been inadequate in addressing these variability and complexities due to their limited capacity for processing extensive demand datasets and reliance on static modeling approaches and simplistic assumptions. In response, there has been a substantial shift towards data-driven methods, particularly those facilitated by machine learning techniques and advanced data analytics, to offer a compelling advantage by providing enhanced predictive accuracy and real-time adaptability \citep{bajari2015machine}. These advanced methods can analyze vast amounts of historical and real-time data, identifying demand patterns and trends that are not readily apparent through traditional methods, which are essential to address fluctuating demand for delivery and meet the evolving demands of modern cities.


Despite great advances that have been made in many real-world problems \citep{zeng2019last,li2021learning,wu2019deepeta,gao2021deep,hess2021real,liang2023poisson}, there are still some critical scenarios that remain underexplored and call for new modeling frameworks. 
Consider the case in which the demand values in operational {regions} are fully recorded in a region within a city. When new {regions} are planned for development in proximity to existing regions, managers endeavor to estimate and predict potential demands, as illustrated in Fig. \ref{fig:intro}. 
Furthermore, when a business seeks to expand to a new city where demand data are recorded over a relatively short period, it is essential to promptly evaluate the projected levels of future demand based on the limited records directly in order to ensure the adaptability of delivery strategies.
In fact, operations in new regions have no historical data for reference. The model needs to exploit patterns from existing regions or transfer knowledge from the source cities to successfully predict expected demands.

To address this challenging issue, we break down the complexity into three key aspects as follows.

\textit{1. How to model intricate interactions between the demand patterns of existing {regions} and newly developed ones.} It is anticipated that the construction of warehouses {in new regions} will influence the demand evolution of existing {regions}. Concurrently, the operational sphere of influence of preexisting warehouses will similarly exert a synergistic effect on the demand for new warehouses. This leads to an intricate interaction between the two items and is crucial for accurate demand estimation and prediction.
{Therefore, we assume that these interactive phenomena are composed of both region-specific patterns and region-wise dependencies, and propose to capture them using individual-collective graph representations.}


\textit{2. How to integrate adequate information to estimate unmeasured demands.} The delivery demand is influenced by many associated factors \citep{fancello2017investigating,fabusuyi2020estimating}. In particular, geospatial factors play an essential role in the emergence of last-mile delivery demand \citep{suguna2021study}. In this sense, collecting sufficient geolocational data becomes a prerequisite for an accurate estimation of city demand. However, collecting such data is nontrivial and requires a laborious process. In addition, different cities can have different data availability and source types. Specific geospatial data collected in a city may not be applicable to another city. An accessible data source is an urgent necessity. Fortunately, utilizing recent advances in large language models (LLMs), we can extract general geospatial knowledge from the unstructured locational data by exploiting the context reasoning ability of them. By integrating the transferable knowledge from LLMs into the spatial-temporal graph neural network (STGNN), our architecture can serve as a generalized demand estimator.


\textit{3. How to guarantee the transferability of the model in new regions and cites.} Last, operators and managers often confront with the ``cold-start'' scenario in delivery operation. Models overfit to the training demand distribution may fail to transfer to new regions or cities without sufficient historical data for model re-training and fine-tuning. Therefore, the zero-shot transferability can facilitate the dynamic adjustment and adaptability of the strategic design. To achieve this, we first construct a functional graph that is generalizable to different cities relying on the encoding of LLMs. Then the backbone STGNN predictor is trained in an inductive manner with a joint reconstruction and forecasting task, ensuring its ability to generalize to new regions and cities in an end-to-end routine. {Since the LLM-generated encoding requires no training and can be obtained offline, it enables prediction with varying numbers of new regions and cities at different scales.}
Based on this strategy, our model achieves three levels of transfer: (1) new regions without any demand records; (2) new cities with full current demand observations; and (3) new cities with partial current demands.


To evaluate the effectiveness of our method, we perform numerical experiments on two real-world delivery order datasets in China and the US. The task includes both joint demand estimation and prediction and cross-city transfer.
Extensive empirical results demonstrate that our model significantly outperforms state-of-the-art baselines in these challenging tasks. Comprehensive discussions are also provided to enhance interpretability.
This paper involves the following contributions:
\begin{itemize}
    \item 
    {We highlight the significance of location-based modeling in delivery demand joint estimation and prediction task and develop a data-driven scheme for extracting geolocation knowledge from LLMs};
    \item {We present a method for integrating geolocation encoding into graph-based deep learning architectures. This approach allows for the modeling of both region-specific patterns and region-wide interactions in a unified graph representation, while simultaneously preserving the transferability across cities.}
    \item Comprehensive evaluations using real-world urban delivery order datasets containing 8 cities from both China and the US demonstrate the effectiveness of our model. {It is demonstrated that the performance gains in accuracy of the encoding are generic and it substantially improves cross-city transferability.}
\end{itemize}

The rest of this paper is organized as follows. Section \ref{sec:related_work} reviews the existing literature on delivery demand management, spatiotemporal data modeling, and LLMs. 
Section \ref{sec:method} first provides the conceptualization and notation of the problem, then formulates the model framework and the solution. 
Section \ref{sec:experiments} evaluates the model in real-world delivery datasets. Section \ref{sec:discussion} provides a detailed discussion of the results and further studies. Section \ref{sec:conclusion} concludes this work and provides future directions.

\section{Related Work}\label{sec:related_work}
In this section, we first review the related literature in urban delivery demand management and survey several representative tasks on this topic. Since the problem under study is related to the spatio-temporal data learning problem, we then briefly introduce recent advances in LLM-based spatio-temporal data modeling and advanced graph neural architectures.

{\subsection{Data-driven Urban Delivery Demand Management}}
The significant increase in urbanization and the advent of online commerce have elevated delivery demand management to a central concern in the realm of urban logistics systems. To evaluate the operational performance of the existing warehouse and delivery strategy, the records of pickup and delivery orders are collected and processed to develop analytical and data-centric models \citep{wen2024survey}. These models are used to address real-world problems associated with logistics demand management, including last-mile delivery route planning \citep{zeng2019last, li2021learning}, time of arrival estimation \citep{wu2019deepeta,gao2021deep}, route prediction \citep{wen2022graph2route}, and demand prediction \citep{hess2021real,liang2023poisson,nuzzolo2014urban}. 
The majority of existing studies focus on package delivery or food delivery processes, utilizing private data within a singular logistics system. This increases the probability of drawing a conclusion that is biased towards a specific dataset.
It is fortunate that an open-source dataset from industry \citep{wu2023lade} considerably facilitates data-driven applications. In addition to the above-studied problems, another fundamental task is to predict the demand level of newly developed {regions}. The issue is further complicated by the lack of historical data on the performance of new {regions}, as well as the impact of existing warehouses on the operations of new businesses.

\subsection{Large Language Models for Spatial-Temporal Data}
In recent years, large language models (LLMs) have made significant strides in the field of natural language processing. The advent of generative pre-trained transformers (GPTs) has ushered in a new era of modern machine learning, where large models have the potential to become a foundational architecture for a range of general language-related tasks. Representative LLMs include BERT \citep{devlin2018bert}, Llama \citep{touvron2023Llama}, and Mistral \citep{jiang2023mistral}, and have been widely adopted for a wide range of disciplines. 
LLMs are distinguished for their excellent reasoning ability and in-context learning capability, which allows for few-shot or even zero-shot tasks \citep{brown2020language,kojima2022large,wei2021finetuned}.
A large number of studies have devoted to explore the potentials of LLMs to address fundamental problems in various areas. Most related to the problem studied, LLMs have demonstrated effectiveness in time series forecasting \citep{jin2023time,garza2023timegpt,jia2024gpt4mts,chang2023llm4ts}, spatial-temporal forecasting \citep{li2024urbangpt,jin2023large,zhang2024large}, and geospatial knowledge mining \citep{gurnee2023language,manvi2023geollm,manvi2024large,tang2024synergizing}. {Moreover, LLMs have shown the potential to be integrated into transportation and mobility research, such as delivery route optimization \citep{liu2023can,qu2023envisioning}, traffic data imputation \citep{zhang2024semantic}, mobility demand prediction \citep{wang2023would,liang2024exploring,gong2024mobility}, individual travel routine generation \citep{wang2024large}.
}
Most of these methods adopt LLMs as backbone predictors either by transforming the input domain data to language tokens or fine-tuning LLMs with domain-specific instructions. In contrast to the aforementioned approach, we utilize LLMs as a generic geospatial encoder and enhance the downstream specialized model through the incorporation of LLM embedding, as opposed to employing them as the fundamental architectural component.

\subsection{Spatial-Temporal Graph Neural Networks}
Spatial-temporal graph neural networks (STGNNs) have become the leading approaches for graph-based spatiotemporal forecasting.
STGNNs adopt sequential techniques such as recurrent neural networks (RNNs), temporal convolutional networks (TCNs) and self-attention to process temporal correlations.
Spatial techniques such as graph convolution, node embedding, and message-passing neural networks are employed for exploiting spatial relational interactions. STGNNs have shown promising results using them for traffic flow forecasting, air quality prediction, and energy consumption management \citep{yu2017spatio, li2018diffusion, wu2019graph, bai2020adaptive, guo2019attention, zheng2020gman, shao2022decoupled,cini2022scalable}.

A recent trend has shifted the focus from the design of advanced architectures to the manipulation of data-specific patterns. Learnable node embedding is a representative solution \citep{STID,cini2023taming, nie2024contextualizing}. Learnable embeddings are fed into modules of STGNNs and learned end-to-end with the forecasting tasks, enhancing the ability to identify node-specific patterns.
Specifically, \cite{shao2022spatial} used learnable embeddings for all nodes, time of day, and day of week stamps, allowing MLPs to achieve competitive performance in several datasets. 
\cite{cini2023taming} offered a systematic interpretation of node embedding as local effects and integrated it into a global-local architectural template. Fourier positional encoding was adopted in \citep{nie2024spatiotemporal} to enable canonical MLPs to capture high-frequency structures in traffic data.

Apart from the spatiotemporal forecasting tasks, STGNNs have also demonstrated effectiveness in other important tasks, such as the time series and spatiotemporal data imputation \citep{liang2022memory,GRIN,marisca2022learning,nie2024imputeformer, xue2024network}, spatiotemporal kriging \citep{wu2021inductive,SATCN,nie2024contextualizing,zheng2023increase}, and graph-based virtue sensing \citep{de2024graph}. 
As we aim to forecast demand patterns for both operating areas with historical records and new areas without histories, the joint demand estimation and forecasting problem studied lies at the intersection of the two research venues, needing an integrated solution.

{Similar to the delivery demand estimation task in urban logistics system, recent studies have elaborated advanced STGNNs to address motility demand estimation problems in both micro and macro transportation systems \citep{xue2025data}. Representative works include bike sharing demand prediction \citep{liang2023cross}, dynamic flow distribution prediction in dockless e-scooter sharing system \citep{he2020dynamic},
trip generation for bike sharing platform planning \citep{liang2023deep,liang2024time}, and mobility flow generation \citep{yao2020spatial,zhang2023deep,liang2024generating}. Despite promising performance, these methods rely on handcrafted geospatial features with an arduous data collection process, such as POIs and distance metrics, to construct underlying graphs used as input to the model. Particularly, the transfer between observed locations and unobserved locations has been achieved in previous work by aggregating the representation learned for observed locations using a prescribed similarity metric \citep{liang2024time,wei2024inductive}. However, this is sensitive to the choice of computation rules and the availability of supplementary data, making the model less generalizable across cities.
}

\section{Model Formulation}\label{sec:method}
\begin{figure}[!htbp]
  \centering
  \captionsetup{skip=1pt}
  \includegraphics[width=1\columnwidth]{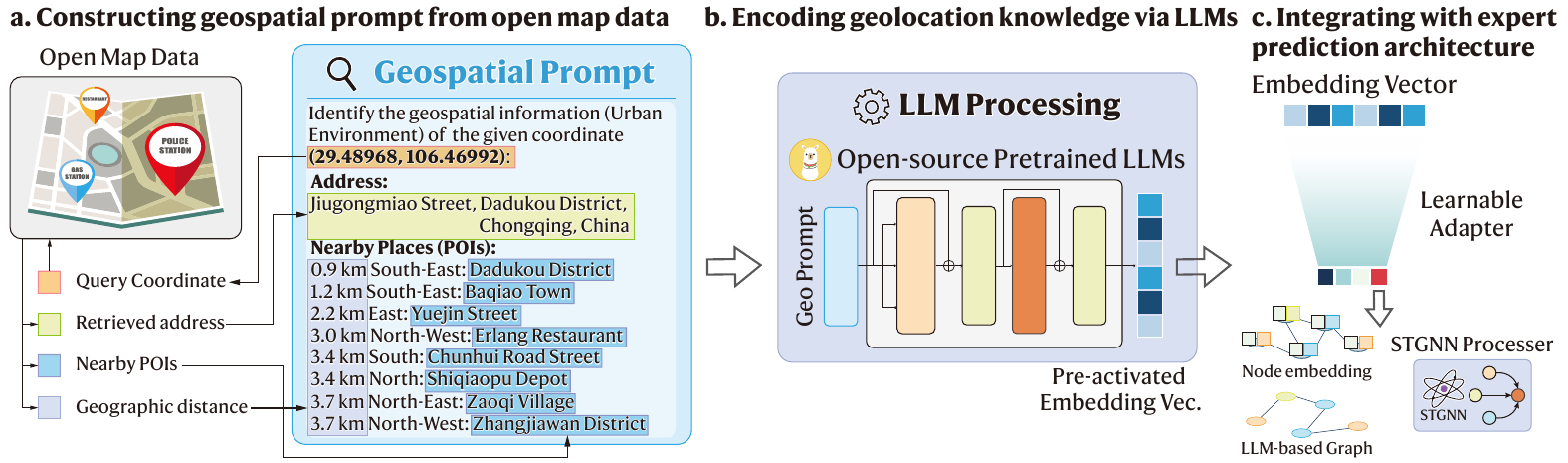}
  \caption{{Overview of the methodology. We query LLMs using a prompt containing a basic geolocation description of the target region retrieved from OpenStreetMap. Then, the pre-activation embedding is extracted from the pre-trained LLM and fed into the downstream STGNN. The STGNN processes the vector in both node embedding and graph construction, enhancing transferability and accuracy.}}
  \label{fig:method}
\end{figure}

In this section, we first formalized the demand joint estimation and prediction problem under the graph-based deep learning framework. Then techniques for extracting geospatial knowledge from LLMs are introduced. Finally, we formulate the model architecture and integrate all elements into an end-to-end inductive training scheme.
The overview of our method is shown in Fig. \ref{fig:method}. For brevity, we name our model: \underline{I}nductive \underline{M}essage-\underline{P}assing Neural Network with \underline{E}ncoding from \underline{L}LMs (\textsc{Impel}).

\subsection{Problem Conceptualization and Formulation}
This paper processes the package and food delivery order datasets.
These data contain the information associated with each order such as the pickup time $\tau_i$ and location $\vv_i=(x,y)_i$. By matching the location of each pickup to the region of interests (ROIs) on the map and aggregating the total number of pickups to a specific time window, a dynamic demand matrix $\mX\in\mathbb{R}^{N\times T}$ can be obtained:
\begin{equation}
    x_t^n=\sum_i d_{n,t}^i, ~d_{n,t}^i=\left\{ \begin{array}{ll}
         1, ~\text{if}~ \tau_i\in I_t \cap (x,y)_i\in R_n \\
         0, ~\text{else} 
    \end{array}\right., \forall n\in\{1,2,\dots,N\}, t\in\{1,2,\dots,T\},
\end{equation}
where $R_n$ and $I_t$ are the $n$-th ROI and $t$-th time interval. This definition can be readily extended to the multivariate case $\vx_t^n\in\mathbb{R}^{d_x}$ that each region has $d_x$-dimensional observations, such as the demand of different products.
Given regions $\{R_n\}_{n=1}^{N_o}$ where operations are already underway, managers expect to take advantage of its historical demand records to predict future demand values, as well as the estimated demand values for newly commissioned warehouses in regions $\{\bar{R}_n\}_{n=1}^{N_u}$.
In addition, each region $R_n$ ($\bar{R}_n$) is associated with a spatial location. We use the central point to represent the coordinate $\{\vv_n\}_{n=1}^{\Bar{N}}$ of each region with $\Bar{N}=N_o+N_u$ and assume that all coordinates are available.

Then, given regions with available observations $\{\mX^n_{\{1,\dots,T\}}\in\mathbb{R}^{T\times D}\}_{n=1}^{N_o}$ and coordinates $\{\vv_n\}_{n=1}^{\Bar{N}}$, for the next time horizon with length $T'$,
the joint objective is to estimate a posterior probability:
{
\begin{equation}\label{eq:objective}
    p(\underbrace{\{\Bar{\mX}^n_{\{1,\dots,T\}}\}_{n=1}^{N_u}}_{\text{unobserved demand}},\underbrace{\{\mX^n_{t+T}\}_{n=1}^{\Bar{N}}}_{\text{future demand}}|\underbrace{\{\mX^n_{\{1,\dots,T\}}\}_{n=1}^{N_o}}_{\text{observed demand}},\{\vv_n\}_{n=1}^{\Bar{N}}),~\forall t\in [1,T'],
\end{equation}
where the unobserved historical demand is estimated and future demands for all regions are predicted. The standard demand time series forecasting task exploits rich historical patterns, and the demand estimation task emphasizes the relations between spatial units. The problem studied lies in the junction of the two.
In contrast to previous work on demand estimation, we also advocate for the \textit{transferability} of the model between different regions and cities, which is of high practical significance and yet to be fully explored.}

As stated above, for new regions, both the historical and future values need to be estimated. For existing regions, the future values are based on both their historical demands and the existence of new regions.
To approximate Eq. \ref{eq:objective}, our solution is to learn a predictive model parameterized by $\vtheta$ :
\begin{equation}
    \vtheta = \arg\max_{\vtheta} \mathbb{E}[p_{\vtheta}(\{\Bar{\mX}^n_{\{1,\dots,T\}}\}_{n=1}^{N_u},\{\mX^n_{t+T}\}_{n=1}^{\Bar{N}}|\{\mX^n_{\{1,\dots,T\}}\}_{n=1}^{N_o},\{\vv_n\}_{n=1}^{\Bar{N}})].
\end{equation}

To learn a parameterized function, we can exploit the relational dependencies existed in the training data as a reference. Three prominent types of dependencies that characterize this task are identified:
\begin{equation}
\begin{aligned}
    \mathcal{D}_{\text{intra}}^o&=\{\text{Corr}(\vx_t^i,\vx_t^j) |~ \forall R_i,R_j \in \{R_n\}_{n=1}^{N_o}\} , \\
    \mathcal{D}_{\text{intra}}^u&=\{\text{Corr}(\bar{\vx}_t^i,\bar{\vx}_t^j) |~ \forall R_i,R_j \in \{\bar{R}_n\}_{n=1}^{N_u}\} , \\
    \mathcal{D}_{\text{inter}}^{o,u}&=\{\text{Corr}(\vx_t^i,\bar{\vx}_t^j) |~ \forall R_i\in \{R_n\}_{n=1}^{N_o},\forall R_j \in \{\bar{R}_n\}_{n=1}^{N_u} \} , \\
\end{aligned}
\end{equation}
which denote the intra-location and inter-location covariate sets, respectively. $\mathcal{D}_{\text{intra}}^o$ denotes the correlation between observed regions, which can be used to learn the relations between observations to predict future values. This is frequently discussed in the time series forecasting setting. $\mathcal{D}_{\text{intra}}^u$ includes the latent interactions between unobserved variables, which can be inferred from the message propagation of latent variables.
Finally, learning in $\mathcal{D}_{\text{inter}}^{o,u}$ is the key to estimating unmeasured demands for new regions. In this case, a similarity metric is needed to describe relations between observed and unobserved locations. For example, proximity between coordinates $\{\vv_n\}_{n=1}^{\Bar{N}}$ can be used as a proxy in kriging models \citep{wu2021inductive}. 
Instead, we model these dependencies using the functional similarity based on the knowledge extracted from LLMs. Meanwhile, we also associate each location with an encoding derived from LLMs, serving as an additional covariate for $\mathcal{D}_{\text{inter}}^{o,u}$.

The main notations used in this work are prescribed in Tab. \ref{tab:my_label}.

\begin{table}[!htbp]
\caption{Main notations used throughout the paper.}
\vspace{0pt}
    \centering
    \begin{small}
    \begin{tabular}{l|p{3.25in}}
    \toprule
    $\displaystyle \vx_t^i$ & The demand record of region $i$ at time step $t$\\
    $\displaystyle \mX_{t:t+T}^i$ & The demand series of region $i$ within a period $t:t+T$\\
    $\displaystyle \mX_{t}$ & The collection of $N_t$ demand series at time stamp $t$ \\ 
    $\displaystyle \mathcal{X}_{t:t+T}$ & All correlated demand series within a period $t:t+T$\\
    $\displaystyle \mA$ & The proximity-based adjacency matrix of all regions \\
    $\displaystyle \mA_t$ & The dynamic adjacency matrix of all regions at step $t$\\
    $\displaystyle \mV$ & The coordinates of all regions  \\
    $\displaystyle \vu_t$ & The time-related exogenous variable \\
    $\displaystyle \vm_{t:t+T}^i$ & An indicating variable to denote whether region $i$ is observed within $t:t+T$ \\ 
    $\displaystyle \gG_t$ & The spatiotemporal graph comprised of several elements \\
    $\displaystyle W, H$ & The historical window size and forecasting horizon \\
    $\displaystyle || \vx ||_p $ & $\normlp$ norm of $\vx$ \\
    $\displaystyle \xi, \sigma, \psi, \rho$ & Nonlinear functions \\
    $\displaystyle f(\vx ; \vtheta) $ & A function of $\vx$ parametrized by $\vtheta$.
    (Sometimes we write $f(\vx)$ and omit the argument $\vtheta$ to lighten notation) \\
    $\displaystyle \mathcal{F}$ & A composition of functions, e.g., a graph neural network \\
    $\displaystyle \mathcal{L}$ & A deterministic loss function \\
    $\displaystyle p(\cdot)$ & A probability distribution over a continuous variable \\
    $\displaystyle \vx \sim p$ & The variable $\ra$ has distribution $P$ \\
    $\displaystyle  \E_{\vx\sim p} [ f(x) ]$ & Expectation of $f(x)$ with respect to $p(\vx)$ \\
    \bottomrule
    \end{tabular}
    \end{small}
    \label{tab:my_label}
\end{table}

\subsection{Joint Demand Estimation and Prediction through Graph Models}
We address the city-wide delivery demand joint estimation and prediction problem by formulating it as a graph-based spatiotemporal learning task. Based on the graph representation model, we then develop an inductive model training procedure to encourage transferability in new cities and regions.

\subsubsection{Modeling Demand Patterns with Spatial-Temporal Graph Representations}\label{sec:graph_representation}

We start by modeling the dynamic demand matrix as a multivariate time series set. Generally, each series $i$ can have a $d_x$-dimensional record at each time step $t$, denoted by $\vx_t^i\in\mathbb{R}^{d_x}$. Matrix $\mX_{t:t+T}^i\in\mathbb{R}^{T\times d_x}$ denotes a single observed series within a given time window $\{t:t+T\}$. For simplicity, we also indicate $\mX_{t}\in\mathbb{R}^{N_o\times d_x}$ as the collection of $N_o$ sensors at time $t$. 
The $N$ correlated demand series can be stacked as a tensor $\mathcal{X}_{t:t+T}=\{\mX_{t:t+T}^i\}_{i=1}^{N_o}\in\mathbb{R}^{N_o\times d_x\times T}$. Some exogenous variables can also be collected, e.g., the date/time identifiers and events, and denoted by $\vu_t\in\mathbb{R}^{d_u}$.

The generation of demand is determined by many interwoven factors. We assume that the system of demand generation constructs a time-invariant spatiotemporal stochastic process:
\begin{equation}\label{eq:st_process}
    \vx_{\tau}^i \sim p^i(\vx_{\tau}^i|\mathcal{X}_{<{\tau}},\mU_{\leq \tau}),~\forall i\in\{1,\dots,N_o\},~\forall \tau\in\{t:t+T\},
\end{equation}
where $p$ is the conditional distribution, and $\mU_{\leq \tau}$ is the stacked static features. 

To instantiate Eq. \ref{eq:st_process}, we assume the regularities of regional demand series follow a locality-aware graph polynomial vector autoregressive system \citep{isufi2019forecasting,cini2023taming}:
\begin{equation}\label{eq:gpvar}
\begin{aligned}
    \mH_t&=\sum_{l=0}^L\sum_{p=1}^P\Psi_{p,l}\mS^{l}[\mX_{t-p}\|\vu_{t-p}], \\
    \mX_t&=\ve\odot\xi(\mH_t) + \eta_t,
\end{aligned}
\end{equation}
where $\Psi\in\mathbb{R}^{P\times L}$ is the collection of model parameters, $P$ is the total number of time lags, $L$ is the total order of graph shifts, $\eta_t\sim \mathcal{N}(0,\sigma^2\mathbb{I})$ is the Gaussian noise, $\xi$ adds the nonlinearity, $\mH_t$ is the hidden state at step $t$, $\ve\in\mathbb{R}^{N_o}$ simulates the region-specific patterns, $\mS^l$ is a graph shift operator derived from the graph Laplacian, e.g., $\mS=\mD^{-1/2}(\mI+\mA)\mD^{-1/2}$.

This graph-level spatiotemporal process delineates the evolution of all objects with regions as nodes. 
However, direct inference of parameters in this model is nontrivial, as it contains unknown latent variables. Therefore, we aim at learning a parameterized model $p_{\vtheta}$ to approximate the unknown conditional probability function. Given a discrete historical window $\{t-W,\dots,t\}, \forall t$ we have:
\begin{equation}\label{eq:parameterized_model}
    p_{\vtheta}(\vx_{t+h}^i|\mathcal{X}_{t-W:t},\mU_{t-W:t+h}) \approx p^i(\vx_{t+h}^i|\mathcal{X}_{<t},\mU_{\leq {t+h}}),~\forall h \in [0,H), ~\forall i\in\{1,\dots,N_o\}.
\end{equation}

Based on the assumption in Eq. \ref{eq:gpvar}, the multivariate demand time series is then abstracted as a \textit{spatiotemporal graph}, with the relational relationships among the regions modeled as edges.
The relational structures can be described by functional relationships between regions.
Since the functional graph is nontrivial to obtain, we can use the adjacency graph to approximate the functional graph. Specifically, the adjacency graph can be calculated from the position of each region and their reciprocal physical proximity \citep{DCRNN}, indicated by ${\mA}\in\mathbb{R}^{{N_o}\times {N_o}}$. In a more general case, the number of ROIs can be time-varying, then the dynamic graph $\mA_t\in\mathbb{R}^{N_o^t\times N_o^t}$ is considered. To model regional-level demand patterns, we assume that geographic information about each region is available and invariant with time. We use $\mV\in\mathbb{R}^{{N}_o\times d_v}$ to represent the geolocation property of all ROIs, e.g., the spatial coordinates. Considering above elements, the observations can be organized as a sequence of spatiotemporal graphs $\{\mathcal{G}_t,\dots,\mathcal{G}_{t+T}\}$ with $\mathcal{G}_t=(\mX_t, \mA_t, \mV)$. Note that a basic assumption in this work is that the functional graph can capture the correlational or relational structures among the demand measurements, and the location of all ROIs are available and fixed at each time step.

After considering the multivariate dependencies among the demand series $\mathcal{X}_{t-W:t}$, the graph relational structures can be encoded in the formulation as an inductive bias for the model architecture:
\begin{equation}
    p_{\vtheta}(\vx_{t+h}^i|\mathcal{G}_{t-W:t},\mU_{t-W:t+h}) \approx p^i(\vx_{t+h}^i|\mathcal{X}_{<t},\mU_{\leq {t+h}},\mV),~\forall h \in [1,H], ~\forall i\in\{1,\dots,N_o\}.
\end{equation}

The above formulation only considers existing regions. After including the new regions, the demand joint estimation and forecasting problem in Eq. \ref{eq:objective} can be rewritten using the graph representation:
\begin{equation}
\begin{aligned}
    \overline{\mathcal{G}}_{\tau} &= (\{\vx_{\tau}^i\}_{i=1}^{N_o}, \bar{\mA_{\tau}}, \bar{\mV}), \\
    {\bar{\mathcal{X}}}^u_{t-W:t},\hat{\mathcal{X}}_{t:t+H} &=\mathcal{F}(\overline{\mathcal{G}}_{t-W:t},\mU_{t-W:t+H}|\vtheta)~~ \text{s.t.}~~\hat{\mathcal{X}}_{t:t+H}\approx\mathbb{E}_p[\mathcal{X}_{t:t+H}], \\
\end{aligned}
\end{equation}
where $\bar{\mA_{\tau}}\in\mathbb{R}^{\bar{N}^t\times \bar{N}^t}$, $\bar{\mV}\in\mathbb{R}^{\bar{N}\times d_v}$,
$\hat{\mathcal{X}}_{t:t+H}=\{\mX_{t:t+H}^i\}_{i=1}^{\bar{N}}$, $\bar{\mathcal{X}}^u_{t-W:t}=\{\Bar{\mX}_{t-W:t}^i\}_{i=1}^{{N_u}}$.

\subsubsection{Parameterization by Space-then-Time Message Passing Network}
After preparation of necessary components, we now formulate the expert backbone model, i.e., the parameterized of a spatial-temporal graph neural network (STGNN) $\mathcal{F}$ instantiated by a message passing mechanism. Then we explain how this architectural template can be adapted to model both region-wise and city-wise demand dynamics.

Specifically, to model the time-varying representation of the delivery demand graph, we consider a spatiotemporal message passing neural network (STMPNN) with the $\ell$-th layer formulated as follows:
\begin{equation}
    \vx_t^{i,\ell+1}=\psi^{\ell}\left(\mX_{\leq t}^{i,\ell}, \textsc{Aggr}_{j\in\mathcal{N}(i)}\{\rho^{\ell}(\mX_{\leq t}^{i,\ell},\mX_{\leq t}^{j,\ell}, e_t^{i\leftarrow j})\}\right),
\end{equation}
where $\psi^{\ell}: \mathbb{R}^{d_{\ell}}\mapsto\mathbb{R}^{d_{\ell+1}}$ and $\rho^{\ell}: \mathbb{R}^{d_{\ell}}\times\mathbb{R}^{d_{\ell}}\mapsto\mathbb{R}^{d_{\ell}}$ are the message updating and message passing functions with possible nonlinearity. $\textsc{Aggr}$ is a permutation invariant aggregation function, e.g., the summation. $\mathcal{N}(i)$ refers to neighbors of node $i$ in the graph and $e_t^{i\leftarrow j}$ is the edge weight at time $t$.

To simplify the process, we assume the relational structure between nodes keeps static, and the above equation can be decoupled and instantiated as a \textit{time-then-space} architecture \citep{MPGRU}:
\begin{equation}\label{eq:time-then-space}
    \vx_t^{i,0} = \textsc{TempEnc}(\mX_{t-W:t}^{i,0}),~ \vx_t^{i,\ell+1}=\psi^{\ell}\left(\vx_{t}^{i,\ell}, \textsc{Aggr}_{j\in\mathcal{N}(i)}\{\rho^{\ell}(\vx_{t}^{i,\ell},\vx_{t}^{j,\ell}, e^{i\leftarrow j})\}\right)_{\ell=0}^{L-1},
\end{equation}
where $\textsc{TempEnc}(\cdot)$ is a temporal processing module that correlates the point observation of each input sequence within the time window $\{t-W,\dots,t\}$. The time-then-space architectural template provides great efficiency benefits compared to alternating spatial and temporal processing. The $\textsc{TempEnc}(\cdot)$ aggregates temporal information at once; then the spatial (multivariate) dependencies are modeled by the spatial aggregation. Eq. \ref{eq:time-then-space} constitutes the backbone architecture of our model.

\subsubsection{Collective-Individual Pattern Treatment Considering Transferability}\label{sec:global_local}

$\mathcal{F}$ defined in above section is trained to make predictions for a union set of all ROIs, without considering any region-specific demand patterns. This \textit{collective} model that conditions on all multivariate demand series globally, explicitly capturing their correlations, and thus allowing transfer in new cities or regions in an inductive manner based on the training scheme in section \ref{sec:inductive_learning}. This makes global models suitable for cold-start scenarios, i.e., demand estimation for newly developed regions. However, it neglects the treatment of region-specific patterns. To consider individual effects and reformulate the global architecture in Eq. \ref{eq:time-then-space}, we first revisit Eq. \ref{eq:parameterized_model} following the terminology in \citet{cini2023graph}.
A global data generating model is given by:
\begin{equation}
    p_{\vtheta}(\vx_{t+h}^i|\mX^i_{t-W:t},\mU_{t-W:t+h},\vv^i) \approx p^i(\vx_{t+h}^i|\mathcal{X}_{<t},\mU_{\leq {t+h}},\mV),~\forall h \in [1,H], ~\forall i\in\{1,\dots,\bar{N}\},
\end{equation}
where $\vtheta$ is the collection of global parameters that can be learned by optimizing the empirical loss over the entire demand series set $\mathcal{X}_{<t}$. Note that the demand variable $\mathcal{X}_{<t}$ contains latent variables, i.e., the demand of unmeasured regions.
To further model the interaction among the demand patterns of different regions, i.e., the impact of local neighbor regions, we consider the multivariate extension:
\begin{equation}
    p_{\vtheta}(\vx_{t+h}^i|\bar{\mathcal{G}}_{t-W:t},\mU_{t-W:t+h}) \approx p(\mathcal{X}_{t+h}|\mathcal{X}_{<t},\mU_{\leq {t+h}},\mV),~\forall h \in [1,H], ~\forall i\in\{1,\dots,\bar{N}\},
\end{equation}
where a joint distribution $p$ of all nodes approximates the union of all individual distribution $p^i$ and $\vtheta$ is shared across all regions. Indeed, if the global model is optimized over each series independently, e.g., the channel-independence \citep{nie2022time}, it approximates the joint distribution as:
\begin{equation}
    p_{\vtheta}(\vx_{t+h}^i|\bar{\mathcal{G}}_{t-W:t},\mU_{t-W:t+h}) \approx \Pi_{i=1}^{\bar{N}} p^i(\vx^i_{t+h}|\mX^i_{<t},\mU_{\leq {t+h}},\vv^i).
\end{equation}

Collective models have larger model capacities to process the relational information between regions. 
Since collective models are learned by optimizing the expectation of the loss function over all regions, they may struggle to model region-specific patterns, i.e., $p^i$.
However, as stated in Eq. \ref{eq:st_process}, each demand series for a region can arise from different stochastic processes. The collective model needs an impractical large capacity to account for these variations (i.e., $\Pi_{i=1}^{\bar{N}} p^i$).

Instead, a node-level individual model used to approximate the data generating process is given by:
\begin{equation}\label{eq:individual_model}
    p^i_{\vtheta_i}(\vx_{t+h}^i|\mX^i_{t-W:t},\mU_{t-W:t+h},\vv^i) \approx p^i(\vx_{t+h}^i|\mathcal{X}_{<t},\mU_{\leq {t+h}},\mV),~\forall h \in [1,H], ~\forall i\in\{1,\dots,\bar{N}\},
\end{equation}
where $\vtheta_i$ is the $i$-th individual parameter fitted on the $i$-th region itself. In this case, each region can have individual model $p^i$ and parameter $\vtheta_i$. Individual models fit each specific demand series with a single model, thus overlooking the potential interaction among regions. Indeed, they are transductive that have difficulty in generalizing to new regions, i.e., the region set is \textit{fixed} during all stages.

In practice, it is important to consider individual effects in the formulation of demand models. For instance, the demand pattern of each ROI is impacted not only by some shared patterns such as the holiday, promotional events, and seasonality, but also by their respective routines such as local events.

An ideal solution is to combine the strength of both collective and individual models \citep{grazzi2021meta,cini2023taming}. Following \citep{cini2023taming}, the key step is to add specialized components to the global model, achieving a balance between the individual and collective modeling perspectives. Specifically, a typical kind of collective-individual hybrid model is given by:
\begin{equation}\label{eq:hybrid_model}
    p^i_{\vtheta,\{\phi_i\}}(\vx_{t+h}^i|\mX^i_{t-W:t},\mU_{t-W:t+h},\vv_{\phi_i}) \approx p^i(\vx_{t+h}^i|\mathcal{X}_{<t},\mU_{\leq {t+h}},\mV), ~\forall i\in\{1,\dots,\bar{N}\},
\end{equation}
where $\{\phi_i\}_{i=1}^{\bar{N}}$ is the set of region-dependent parameters and we adopt it to parameterize each individual an index-based component, i.e., $\vv_{\phi_i}\in\mathbb{R}^{d_{\text{emb}}}$. $\vtheta$ is the region-independent parameter shared globally. 
It is straightforward to find that the hypothesis class of all individual models in Eq. \ref{eq:individual_model} is $\mathcal{H}_{\text{Ind}}=\{p(\cdot;\vtheta_i)|\vtheta_i\in\Theta,i=1,\dots,\bar{N}\}$ where $\Theta$ is the parameter space. However, such a hypothesis class is too expensive and computationally infeasible for a large number of instances, e.g., a large dimension of regions. Instead, the reduced hypothesis class of collective-individual model is $\mathcal{H}_{\text{hybrid}}=\{p(\cdot;\vtheta,\phi_{i})|\vtheta\in\Theta,\vv_{\phi_i}\in\mathbb{R}^{d_{\text{emb}}}\}$, which is more feasible and efficient in optimization.

Finally, to implement Eq. \ref{eq:hybrid_model} in a parameter-efficient way, \cite{shao2022spatial} and \cite{cini2023taming} propose to assign each series a learnable embedding that can learn individual-dependent parameters during the training of the backbone architecture. Such a specialization provides flexibility for the globally shared model to condition representations at each series. Specifically, the predictor $\widehat{\mathcal{X}}_{t:t+H}=\mathcal{F}(\bar{\mathcal{G}}_{t-W:t},\mU_{t-W:t+H}|\vtheta;\{\phi_i\}_{i=1}^{\bar{N}})$ with a shared architecture $\mathcal{F}$ can be specialized by modulating the input and output layers using $\vv_{\phi_i}\in\mathbb{R}^{d_{\text{emb}}}$ as follows. 

\begin{equation}\label{eq:stgnn}
\begin{aligned}
    &\vh_t^{i,(0)}=\textsc{TempEnc}(\mX_{t-W:t}^{i,(0)},\mU_{t-W:t},\vv_{\phi_i}), \\
    &\vh_t^{i,(\ell+1)}=\psi^{(\ell)}\left(\vh_{t}^{i,(\ell)}, \textsc{Aggr}_{j\in\mathcal{N}(i)}\{\rho^{(\ell)}(\vh_{t}^{i,(\ell)},\vh_{t}^{j,(\ell)}, e^{i\leftarrow j})\}\right)_{\ell=0}^{L-1}, \\
    &\hat{\vx}_{t+h}^{i}=\textsc{Output}(\vh_t^{i,(L)},\mU_{t:t+h},\vv_{\phi_i}),  ~\forall h \in [1,H], \\
\end{aligned}
\end{equation}
where $\textsc{Output}$ is usually implemented by a single MLP, and $\textsc{TempEnc}$, $\psi$, $\rho$, and $\textsc{Output}$ are shared by all regions. Instead of learning individual models for all regions in Eq. \ref{eq:individual_model}, the above scheme is more parameter-efficient and reduces the risk of overfitting. Before solving for this learning model, we need to address some queries: (1) The demand variable $\mathcal{X}$ has latent variables, i.e., the demand of unmeasured regions; (2) Introducing region-specific parameters to the model that is not applicable to other regions sacrifices the transferability provided by pure collective models; (3) The characterization of geolocation information $\mV$ of all regions requires the integration of sufficient spatial data related with logistics demand, e.g., positions of interests.

\section{Solution Framework}

\begin{figure}[!htbp]
  \centering
  \captionsetup{skip=1pt}
  \includegraphics[width=1\columnwidth]{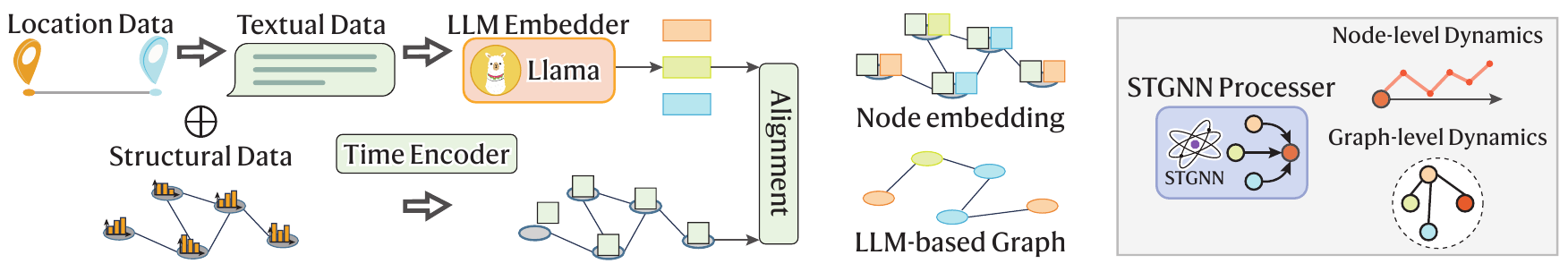}
  \caption{Overview of the solution framework.}
  \label{fig:solution}
\end{figure}

Although the above formulation can characterize variations and correlations of demand patterns at both levels, it is nontrivial to obtain such a parameterization to approximate Eq. \ref{eq:gpvar}. Specifically, the technical challenges for the model solution are identified in three aspects. First, how to capture the node-level dynamics and graph-level interactions simultaneously and efficiently. Second, how to allow the model to transfer to new regions and new cities without retraining. Third, how to integrate essential geolocation knowledge into the model that is generalizable to different cities.
To address these challenges, this section presents our solution framework that seamlessly aligns the embedding of LLMs with GNNs. This LLM-empowered graph-based learning solution provides a learnable and feasible parameterization to approximate Eq. \ref{eq:gpvar}. The solution framework is shown in Fig. \ref{fig:solution}.

\subsection{Encoding Geospatial Knowledge from Large Language Models}
Based on the above descriptions, this section introduces the integration of LLMs to encode geospatial knowledge and graph-based relational bias into the backbone architecture, i.e., the joint demand estimator and predictor. The role of LLM-based encoding is two-fold: (1) transferability region-specific embedding; (2) functional graph based on the similarity of geospatial encoding vector.

\subsubsection{Location-based Geospatial Query}
The sensory data such as demand series is highly associated with some location-dependent variables.
However, collecting sufficient geospatial information, including points of interest, road network geometry, and buildings, can be a laborious and tedious process. This is especially difficult when different cities have different sources and availability of location data.
Fortunately, recent advances in natural language processing provide an alternative paradigm for mining geospatial knowledge \citep{gurnee2023language,manvi2024large}. As LLMs have shown an impressive ability to reason about geospatial data \citep{manvi2023geollm}, it is expected to extract general location-based representations based on the intrinsic knowledge of LLMs.

To achieve this, we first construct a location-based query and organize a text prompt as input to LLMs. 
First, given the query coordinate $\vv=(x,y)$ of the target location, a public map service or database such as OpenStreetMap (OSM) \citep{haklay2008openstreetmap} is accessed and some general geolocation information is obtained, such as POIs and surrounding streets and buildings.
Then a text message containing the basic location information extracted from the OSM is generated as a prompt and fed into LLMs, as shown in Fig. \ref{fig:method}. LLMs are requested to identify and understand the geolocation information, relying on their ability to reason in context.

\subsubsection{Prompt-based Geospatial Reasoning and Encoding Generation}
Different from existing frameworks that adopt LLMs as the backbone forecasting backbone \citep{jin2023time}, or directly generate geospatial message from the output \citep{manvi2023geollm}, we propose to adopt the pre-activated embedding from the output layer of a pre-trained LLM as the latent covariate, which is assumed to be geospatially encoded by the network processing.
To extract the general geospatial encoding from LLMs that is applicable for downstream tasks, we use the linear probing technique \citep{alain2016understanding,belinkov2022probing}. Specifically, given the network activation of hidden states (before the last \texttt{softmax} function) $\mH_{\text{LLM}}\in\mathbb{R}^{\bar{N}\times D_{\text{LLM}}}$ of LLMs, we can use a linear ridge regressor as the probe to fit the node embedding of each ROI:
\begin{equation}\label{eq:ridge_regressor}
    \widehat{\mW}_{\text{prob}} = \arg\min_{\mW_{\text{prob}}}\|\mE-\mH_{\text{LLM}}\mW_{\text{prob}} \|_F^2 + \lambda \| \mW_{\text{prob}} \|_F^2=(\mH_{\text{LLM}}^{\mathsf{T}}\mH_{\text{LLM}}+\lambda\mI)^{-1}\mH_{\text{LLM}}^{\mathsf{T}}\mE,
\end{equation}
where $\mE\in\mathbb{R}^{\bar{N}\times D_{\text{node}}}$ is the empirical node embedding derived from the training data and $\mW_{\text{prob}}\in\mathbb{R}^{D_{\text{LLM}}\times D_{\text{node}}}$. Although $\mE$ can be obtained by learning from the training set or be elaborated by handcrafted rules. However, an ideal node embedding is nontrivial to design across multiple cities with different demand patterns.
Thus, we integrate this process in the end-to-end learning of network parameters (refer to \ref{subsec:llm_integration}). Note that $\widehat{\mW}_{\text{prob}}$ is used to align the task-independent embedding of LLMs with the node representation, making it generalizable for adapting to demand patterns across cities.

\subsubsection{Functional Graph Construction with LLM Encoding}
As discussed in section \ref{sec:graph_representation}, the correlation and interaction between regions are described by a graph. Due to the absence of a precise metric to construct such a graph, physical proximity is generally adopted \citep{lei2022bayesian,nie2023correlating}. However, proximity in the Euclidean is inadequate to reflect the relationships of demand patterns in feature space. Fortunately, the inherent geospatial knowledge in the LLM-based encoding provides an abstract of the geolocation property of each region in the latent space. We can measure the similarity of these encodings and treat it as a more accurate surrogate of the functional similarity.
To achieve this, the functional graph is learned end-to-end simultaneously with the main architecture:
\begin{equation}
    {\bar{\mathcal{X}}}^u_{t-W:t},\hat{\mathcal{X}}_{t:t+H} =\mathcal{F}(\overline{\mathcal{G}}_{t-W:t},\mU_{t-W:t+H},\mA_{\text{LLM}}|\vtheta), ~\mA_{\text{LLM}}\sim q_{\Phi}(\bar{\mA}|\bar{\mathcal{X}}_{t-W:t})\in\mathbb{R}^{\bar{N}\times \bar{N}},
\end{equation}
where $\vtheta,\Phi$ are the base and the graph learning parameter respectively, $\mA_{\text{LLM}}$ is the LLM-based functional graph. To jointly learn the parameters over the training set, we solve the following problem:
\begin{equation}\label{eq:graph learning}
\begin{aligned}
    \vtheta^{\star},\Phi^{\star}&=\arg\min_{\vtheta,\Phi}\frac{1}{T_{\text{train}}}\sum_{t=1}^{T_{\text{train}}}\mathbb{E}_{\mA_{\text{LLM}}\sim q_{\Phi}}[\|\mathcal{F}(\overline{\mathcal{G}}_{t-W:t},\mU_{t-W:t+H},\mA_{\text{LLM}}|\vtheta)-\vx_{t+h}^i\|_2^2], \\
    &=\arg\min_{\vtheta,\Phi}\frac{1}{T_{\text{train}}}\sum_{t=1}^{T_{\text{train}}}\frac{1}{NH}\sum_{i=1}^N\sum_{h=1}^H\|\mathcal{F}(\overline{\mathcal{G}}_{t-W:t},\mU_{t-W:t+H},\mA_{\text{LLM}}|\vtheta)-\vx_{t+h}^i\|_2^2,
\end{aligned}
\end{equation}
where $T_{\text{train}}$ is the total time steps of the training dataset. Eq. \ref{eq:graph learning} is achieved in the forecasting loss in Eq. \ref{eq:inductive learning} and can be solved by typical gradient-based methods.
The following paragraphs will elaborate the backbone architecture $\mathcal{F}$ for processing the graph representation of the demand time series.

\subsection{LLM-Enhanced Collective-Individual Graph-based Learning}\label{subsec:llm_integration}
Next, we demonstrate how the encoding from LLMs and the LLM-based functional graph can be integrated into the architectural template and elaborate the detailed choice of modular components.

\subsubsection{Integration of Encodings from LLMs}

Recall that applying region-specific parameters in the collective model can efficiently account for the effects that characterize individual treatments, thereby improving the predictive performance on the current task. However, this scheme also sacrifices the flexibility provided by pure collective models: it cannot perform inductive learning in a model-based manner, leading to a compromise in transferability. This is because the learned individual parameter cannot be applied to regions outside the training data.
To compensate for this ideal property, we proposed resorting to generalizable embedding from LLMs. We suggest the integration of the LLM-based embedding in two ways.

First, since LLMs can encode comprehensive knowledge that humans can understand, we exploit this encoding as the co-variate of demand series and incorporate it in both the temporal encoder and message passing modules to serve as the local effects discussed in subsection \ref{sec:global_local}:
\begin{equation}
\begin{aligned}
    &\vv_{\phi_i}=\mH_{\text{LLM}}\widehat{\mW}_{\text{prob}}, \\
    &\widetilde{\vv}_{\phi_i} = \textsc{BatchNorm}(\textsc{LeakyReLU}(\vv_{\phi_i})), \\
    &\vh_t^{i,(0)}=\textsc{TempEnc}([\mX_{t-W:t}^{i,(0)}\|\widetilde{\vv}_{\phi_i}],\mU_{t-W:t}), \\
    &\vh_t^{i,(\ell+1)}=\psi^{(\ell)}\left(\vh_{t}^{i,(\ell)}, \textsc{Aggr}_{j\in\mathcal{N}_k(i)}\{\rho^{(\ell)}([\vh_{t}^{i,(\ell)}\|\widetilde{\vv}_{\phi_i}],\vh_{t}^{j,(\ell)}, e^{i\leftarrow j})\}\right)_{\ell=0}^{L-1}, \\
\end{aligned}
\end{equation}
where $\mH_{\text{LLM}}$, $\widehat{\mW}_{\text{prob}}$ are introduced in Eq. \ref{eq:ridge_regressor}, and $\widetilde{\vv}_{\phi_i}\in\mathbb{R}^{D_{\text{node}}}$ is the LLM-based node embedding, $\textsc{BatchNorm}$ is used to reduce the variance of different encodings, and $[\cdot \| \cdot]$ is the concatenation.

Second, learning an appropriate graph in a data-dependent way (see Eq. \ref{eq:graph learning}) is not easy. To efficiently implement this scheme, we make $\mA_{\text{LLM}}$ \textit{implicitly} conditions on $\bar{\mathcal{X}}_{t-W:t}$. Specifically, we first set a layerwise \textit{adapter} $\mD\in\mathbb{R}^{D^{(\ell)}_{\text{graph}} \times D_{\text{node} }}$ shared by all regions, to reduce the parameter space:
\begin{equation}
    \vv_{g,\phi_i}=\textsc{LeakyReLU}(\mD^{(\ell)}\vv_{\phi_i}),
\end{equation}
where $\Phi=\{\mD^{(\ell)}\}_{\ell=0}^L$ is the parameter set for graph learning in Eq. \ref{eq:graph learning}. Then the LLM-based functional graph is simplified as an undirected graph as follows:
\begin{equation}\label{eq:llm_graph}
    e^{i\rightleftarrows j}=\langle\vv_{g,\phi_i},\vv_{g,\phi_j}\rangle,~\mathcal{E}=\{e^{i\rightleftarrows j}|\forall i,j\in1,\dots,\bar{N}\},
\end{equation}
where $\langle\cdot\rangle$ is the vector inner product, $\mathcal{E}$ is the set of fully-connected graph edges.

Intuitively, for regions without historical time series records, the LLM-based encoding provides an expressive latent variable to regress the target value. For regions with observations, both the data and the encoding contribute to the estimation and forecasting. The aggregation of the neighbor encoding and node value in the LLM-based functional graph enables the model to be inductive.

\begin{remark}(Comparison with Learnable Region (Node) Embedding.)
    Note that both the learnable embedding \citep{STID} and our LLM-based encoding for each region (node) can account for the region-specific patterns. However, the learnable embedding is fitted to the training data distribution with full observations available. When transfers to new regions without available observations for training or fine-tuning, the learnable embedding fails to adapt to new data. Instead, our LLM-based encoding is precomputed prior to model training and fixed throughout the learning process. This makes it generalizable across new regions and cities.
\end{remark}

\subsubsection{Implementation Details of Modular Computations}
Finally, we explain the details of model implementations, particularly the instantiating of Eq. \ref{eq:time-then-space}.
\textbf{Temporal message-passing layers (\textsc{TempEnc})}: 
\begin{equation}\label{eq:tempenc}
    \vh_t^{i,\ell+1}=\psi^{\ell}\left(\vh_{t}^{i,\ell},  {\textsc{Aggr}}_{\leq t}\rho^{\ell}(\vx_{t}^{i,\ell},\mX_{\leq t}^{i,\ell})\}\right).
\end{equation}
In practice, recurrent neural networks (RNNs), temporal convolutional networks (TCN), and Transformers can be adopted for Eq. \ref{eq:tempenc}. As an alternative, MLP-based architectures such as TSMixers \citep{chen2023tsmixer} have proven to be a more efficient and flexible choice. Specifically, we choose:
\begin{equation}\label{eq:temporal_enc_final}
    \vh_t^{i,\ell+1}=\sigma(\mW_t^{\ell}[\vx_t^{i,\ell}\|\vx_{t-1}^{j,\ell}\|\dots\|\vx_{t-W}^{j,\ell}]+\vb_t^{\ell}).
\end{equation}

\textbf{Spatial message-passing layers}: we instantiate it as anistropic graph neural networks (GNNs) \citep{satorras2022multivariate}, with a $k$-th-order extension. For each $k$-hop neighborhood, we have:
\begin{equation}\label{eq:mpnn_final}
\begin{aligned}
    &\text{Message Updating:}~ \vm_t^{j\rightarrow i, \ell}=\sigma(\mW_m^{\ell}[\vh_t^{i,\ell}\|\vh_t^{j,\ell}\|e_t^{i\leftarrow j}]+\vb_m^{\ell}), \\
    &\text{Edge Updating:}~ \alpha_t^{j\rightarrow i, \ell}=\sigma(\mW_e^{\ell}\vm_t^{j\rightarrow i, \ell}+\vb_e^{\ell}), \\
    &\text{Node Updating:}~ \vh_t^{i,\ell+1}=\sigma(\mW_n^{\ell}\vh_t^{i,\ell}+\textsc{Mean}_{j\in\mathcal{N}_k(i)}\{\alpha_t^{j\rightarrow i, \ell}\vm_t^{j\rightarrow i, \ell}\})\\
\end{aligned}
\end{equation}
where $\mW_m\in\mathbb{R}^{d_{\ell}\times(2d_{\ell}+1)},\vb_m^{\ell}, \mW_e\in\mathbb{R}^{1\times d_{\ell}},\vb_e^{\ell}, \mW_n\in\mathbb{R}^{d_{\ell+1}\times d_{\ell}}$ are learnable parameters, and $\mathcal{N}_k(i)$ denotes the $k$-th order neighbors of node $i$ in the graph. When the optional adjacency-based graph is activated, an additional diffusion convolution operator \citep{DCRNN} is further considered:
\begin{equation}\label{eq:diff_conv}
   \vh_t^{i,\ell+1}=\psi^{\ell}\left(\vh_{t}^{i,\ell}, \textsc{Aggr}_{j\in\mathcal{N}(i)}\{\rho^{\ell}(\vh_{t}^{i,\ell},\vh_{t}^{j,\ell}, e^{i\leftarrow j})\}\right)=\sigma(\widetilde{e}^{i\leftarrow j}_k\vh_{t}^{j,\ell}\Theta_{1,k}^{\ell}+\vh_{t}^{i,\ell}\Theta_{2}^{\ell}),
\end{equation}
where $\widetilde{e}^{i\leftarrow j}_k$ is the normalized edge weight for the $k$-th hop, $\Theta_{1,k}^{\ell},\Theta_{2}^{\ell}$ are learnable parameters.

\textbf{Dense Feedforward Layer}
After the temporal and spatial processing, we employ a series of dense feedforward layers to increase the model capacity. Specifically, we adopt a series of residual MLP:
\begin{equation}
     \vh_t^{i,\ell+1}=\sigma(\mW_r^{\ell}\vh_t^{i,\ell}+\vb_r^{\ell})+\vh_t^{i,\ell}.
\end{equation}

\textbf{MLP-based Multistep Output Layer}
Finally, we directly generate the multistep forecast results by using a shared MLP readout layer. $\forall h\in\{1,\dots, H\}$, we have:
\begin{equation}
    \hat{\vx}_{t+h}^{i}=\textsc{MLP}(\vh_t^{i,(L)},\mU_{t:t+h},\vv_{\phi_i}), ~\forall i\in\{1,\dots,\bar{N}\}. \\
\end{equation}

\subsection{End-to-End Transferable Predictor by Inductive Learning}\label{sec:inductive_learning}

\begin{figure}[!htbp]
  \centering
  \captionsetup{skip=1pt}
  \includegraphics[width=1\columnwidth]{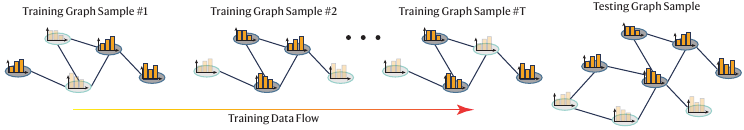}
  \caption{The inductive training scheme. During training stage, the observed {regions} are processed as subgraphs and are randomly masked for reconstruction purposes. During testing stage, the new regions are incorporated according to the prescribed graph structure and locations.}
  \label{fig:training}
\end{figure}

Using the model defined above, a straightforward solution consists of adopting an estimation model to first estimate the demand for unobserved regions and then performing downstream forecasting processing. However, we argue that the future dynamics of demand evolution in existing regions (or depots) can be substantially influenced by the newly developed regions, showing a synchronous effect for their individual demand behaviors. 
Therefore, we propose bypassing the estimation step and considering the forecasting backbone that can directly handle irregularly structured graph data. We aim to jointly estimate demand of newly developed regions and predict the future demand values of already existing {regions} in an end-to-end way, benefiting from the optimization of the shared target.

This can be modeled as an inductive learning task \citep{wu2021inductive,SATCN,nie2023towards} associated with a \textit{kriging} problem \citep{appleby2020kriging,lei2022bayesian,nie2023correlating}.
We first introduce an indicating variable $\vm_{t:t+T}^i\in\{\mathbf{0},\mathbf{1}\}^T$ to denote whether the region $i$ has records or is masked for training purposes. The inductive training scheme constructs a random masking and reconstruction task using the observed data. We then consider the following conditional probability during training:
\begin{equation}
    p(\vm_{t:t+T}^i|\mM_{t:t+T})=p(\vm_{t:t+T}^i)=\mathcal{B}(\beta),~\forall i \in\{1,\dots,N\},~t\in\{1,\dots,T\},
\end{equation}
where $p$ is in fact a random missing probability followed a Bernoulli distribution with the parameter $\beta$ \citep{yi2016st}. Notably, the existence probability of node $i$ is independent of other nodes. Since during the model training, the location and distribution of new regions in the testing stage is unknown for the model. Thus, we do not make any assumptions on the distributions of new regions. In fact, the model should adapt to this distribution shift. For example, in the inference stage, the conditional distribution of new regions can be:
\begin{equation}
    p(\vm_{t:t+T}^i|\{\vm_{t:t+T}^j\}_{j\in\mathcal{N}(i)})\neq p(\vm_{t:t+T}^i),
\end{equation}
which indicates that the observed probability of region $i$ dependents on its neighborhood regions. This is intuitive in urban logistics system. For instance, the newly developed business regions can happen within a nearby region and be influenced by each other.

Based on above discussions, the model needs to address the forecasting problem with irregular graph structures, i.e., the testing graph and node locations can differ from the ones in training. We approach this by applying the masking-reconstruction scheme proposed in \citep{wu2021inductive}. As illustrated in Fig. \ref{fig:training}, this process includes random masking and reconstruction during the training stage.
Consider that we can split the training data into context-target window pairs along the temporal dimension, the following scheme is established. $\forall t\in\{1,\dots,T_{\text{train}}\}$, we have:
\begin{equation}\label{eq:inductive learning}
\begin{aligned}
\cline{0-1}
    &\text{Sampling:}~\vm_{t-W:t}^i \sim p(\vm_{t-W:t}^i)=\mathcal{B}(\beta),~\forall i \in\{1,\dots,N_o\}, \\
    &\text{Masking:}~\Bar{\mathcal{X}}_{t-W:t}=\{\vm_{t-W:t}^i\odot{\mX}^i_{t-W:t}\}_{i=1}^{N_o}, \\
    &\text{Reconstructing:}~\hat{\mathcal{X}}_{t:t+H}=\mathcal{F}(\Bar{\mathcal{X}}_{t-W:t}, \{\mA_{\tau}\}_{t-W}^t|\vtheta)\approx\mathbb{E}_p[p_{\vtheta}(\vx_{t+h}^i|\Bar{\mathcal{X}}_{t-W:t})], ~\forall h \in [1,H]\\
    &\text{Learning:}~\vtheta^{\star}=\arg\min_{\vtheta}\mathcal{L}_{\text{recon}}(\widehat{\mathcal{X}}_{t-W:t},{\mathcal{X}}_{t-W:t})+\mathcal{L}_{\text{pred}}(\widehat{\mathcal{X}}_{t:t+H},{\mathcal{X}}_{t:t+H}) \\
    &=\arg\min_{\vtheta}\underbrace{\sum_{\tau=t-W}^{t}  \frac{\sum_{i=1}^{N_o} \overline{m}_{\tau}^i\cdot \ell(\hat{\vx}_{\tau}^i,\vx_{\tau}^i)}{\sum_{i=1}^{N_o} \overline{m}_{\tau}^i}}_{\text{reconstruction loss}} + \underbrace{\frac{1}{\bar{N} H}\sum_{h=1}^{H}\sum_{i=1}^{\bar{N}}\ell(\hat{\vx}_{t+h}^i,\vx_{t+h}^i)}_{\text{prediction loss}},\\
    \cline{0-1}
    &\text{Estimating:}~\hat{\mathcal{X}}_{t:t+H'}=\mathcal{F}(\bar{\mathcal{X}}_{t-W':t},\{\bar{\mA}_{\tau}\}_{t-W'}^t|\vtheta^{\star}),~\bar{\mathcal{X}}_{t-W':t}=\{{\mX}^i_{t-W':t}\}_{i=1}^{N_o}\cup\{\bar{\mX}^i_{t-W':t}\}_{i=1}^{N_u}. \\
    \cline{0-1}
\end{aligned}
\end{equation}
where $W$ is the window size, $H$ is the forecasting horizon, $\ell(\cdot)$ is some metrics such as the $\ell_1$ norm and mean square error, $\overline{m}_t^i$ is the logical binary complement of $m_t^i$, $\mathcal{L}$ is the inductive training loss function, and the static covariates are omitted for ease of presentation. $\mathcal{L}$ contains two parts: (1) the reconstruction loss in the historical window to estimate historical demand values by exploiting $\mathcal{D}_{\text{inter}}^{o,u}$; (2) the forecast loss in the future horizon to predict future demands by exploiting $\mathcal{D}_{\text{intra}}^o$ and $\mathcal{D}_{\text{intra}}^u$.

Please note that this inductive training scheme is slightly different from the self-supervised masked training scheme used in advanced Transformer-based architectures \citep{du2022saits,nie2024imputeformer,qiu2024routesformer}. The masked training strategy used in sequential models intentionally masked out some consecutive tokens or patches, aiming to learn representations that can benefit downstream tasks in an autoregressive fashion. However, the training scheme designed in this paper is directly associated with the reconstruction task, without assumptions on the missing patterns in the testing set.

Finally, the overall solution framework is summarized in algorithm \ref{algorithm}.

\begin{algorithm}[!htb]
  \setstretch{1.}
  \caption{\textsc{IMPEL} - Overall solution framework.}\label{algorithm}
  \begin{footnotesize}
  \begin{algorithmic}[1]
  \Require  
  Partially observed lookback demand series $\mathcal{X}_{t-W:t}\in\mathbb{R}^{N_o\times d_{x}\times T}$; input length $W$; prediction horizon $H$; total number of ROIs with $\bar{N}=N_o+N_u$; coordinates of all ROIs $\bar{\mV}\in\mathbb{R}^{\bar{N}\times d_v}$; sampling rates $\beta$, hidden dimension $D$; STMP block number $L$; feedforward block number $L_{\text{ffn}}$; adjacency graph $\bar{\mA}\in\mathbb{R}^{\bar{N}\times \bar{N}}$; embedding from LLMs $\mH_{\text{LLM}}$; covariates $\mU_{t-W:t+H}$; learnable model parameters $\vtheta$.
    \State $\triangleright \ $ ------------------------------------Inductive model training------------------------------------ $\triangleleft \ $
    \State $\triangleright \ $ Masking random regions for reconstruction.
    \State $~\vm_{t-W:t}^i \sim p(\vm_{t-W:t}^i)=\mathcal{B}(\beta),~\forall i \in\{1,\dots,N_o\}$, \Comment{$\vm_{t:t+T}^i\in\{\mathbf{0},\mathbf{1}\}^T$.}
    \State $\Bar{\mathcal{X}}_{t-W:t}=\{\vm_{t-W:t}^i\odot{\mX}^i_{t-W:t}\}_{i=1}^{N_o}$,
    
    \State $\triangleright \ $ Activate the LLM-based encoding.
    \State $\vv_{\phi_i}=\mH_{\text{LLM}}\widehat{\mW}_{\text{prob}}, \widetilde{\vv}_{\phi_i} = \textsc{BatchNorm}(\textsc{LeakyReLU}(\vv_{\phi_i}))$, \Comment{$\widetilde{\vv}_{\phi_i}\in\mathbb{R}^{D_{\text{node}}}$.}
    \State $\triangleright \ $ Temporal encoding with LLM-based encoding using Eq. \ref{eq:temporal_enc_final}
    \State $\vh_t^{i,(0)}=\textsc{TempEnc}([\Bar{\mX}_{t-W:t}^{i,(0)}\|\widetilde{\vv}_{\phi_i}],\mU_{t-W:t}),~\forall i \in\{1,\dots,N_o\}$, \Comment{$\vh_t^{i,(0)}\in\mathbb{R}^{D_{\text{emb}}}$.}

    \State $\triangleright \ $ Construct the LLM-based functional graph using Eq. \ref{eq:llm_graph}.
    \State $e^{i\rightleftarrows j}=\langle\vv_{g,\phi_i},\vv_{g,\phi_j}\rangle,~\mathcal{E}=\{e^{i\rightleftarrows j}|\forall i,j\in1,\dots,\bar{N}\},$
    \State $\textbf{for}\ \ell\ \textbf{in}\ \{1,\cdots,L\}\textbf{:}$\Comment{Run STMP blocks.}
    
    \State $\textbf{\textcolor{white}{for}}$ $\triangleright \ $Graph processing using Eqs. \ref{eq:mpnn_final} and \ref{eq:diff_conv} in both the adjacency and LLM-based graphs.
    \State $\textbf{\textcolor{white}{for}}\ \vh_t^{i,(\ell+1)}=\psi^{(\ell)}\left(\vh_{t}^{i,(\ell)}, \textsc{Aggr}_{j\in\mathcal{N}_k(i)}\{\rho^{(\ell)}([\vh_{t}^{i,(\ell)}\|\widetilde{\vv}_{\phi_i}],\vh_{t}^{j,(\ell)}, a^{i\leftarrow j},e^{i\rightleftarrows j})\}\right)_{\ell=0}^{L-1}$,
    \State $\triangleright \ $ Dense feed-forwarding layers and the readout layer. 
    \State $\textbf{for}\ \ell\ \textbf{in}\ \{1,\cdots,L_{\text{mlp}}\}\textbf{:}$\Comment{Run MLP blocks.}
    \State $\textbf{\textcolor{white}{for}}\ \vh_t^{i,\ell+1}=\sigma(\mW_r^{\ell}\vh_t^{i,\ell}+\vb_r^{\ell})+\vh_t^{i,\ell},$
    \State $\hat{\vx}_{t+h}^{i}=\textsc{MLP}(\vh_t^{i,(L_{\text{mlp}})},\mU_{t:t+h},\vv_{\phi_i}), ~\forall i\in\{1,\dots,N_o\}, \ h\in\{1,\dots,H\},$ \Comment{Generate predicted values.}

    \State $\vtheta^{\star}=\arg\min_{\vtheta}\mathcal{L}_{\text{recon}}(\widehat{\mathcal{X}}_{t-W:t},{\mathcal{X}}_{t-W:t})+\mathcal{L}_{\text{pred}}(\widehat{\mathcal{X}}_{t:t+H},{\mathcal{X}}_{t:t+H})$, \Comment{Loss backward.}
    \State \textbf{Return}\ Trained model $\mathcal{F}(\cdot|\vtheta^{\star})$.

    \State $\triangleright \ $ ---------------------------------------------Inference--------------------------------------------- $\triangleleft \ $
    \State $\triangleright \ $ Given new regions with coordinates available,
    \State $\hat{\mathcal{X}}_{t:t+H'}=\mathcal{F}(\bar{\mathcal{X}}_{t-W':t},\bar{\mA},\bar{\mV},\mU|\vtheta^{\star}),~\bar{\mathcal{X}}_{t-W':t}=\{{\mX}^i_{t-W':t}\}_{i=1}^{N_o}\cup\{\bar{\mX}^i_{t-W':t}\}_{i=1}^{N_u}.$
    \State $\textbf{Return}\ \hat{\mathcal{X}}_{t:t+H'}$ \Comment{Return the prediction result.}
  \end{algorithmic} 
  \end{footnotesize}
\end{algorithm}




\section{Experiment}
\label{sec:experiments}
This section offers extensive experiments and in-depth discussions to evaluate the proposed model. We first collect and process two real-world demand datasets, including package delivery data in China and food delivery data in the US.
Our model is compared with state-of-the-art baseline models in the literature, with respect to both prediction accuracy and transferability. Then, we conduct model analysis to highlight the rationality of our model. Finally, case studies provide intuitive illustrations on the prediction results and explanations of LLM-based encodings.

\subsection{Dataset Description}

Two real-world logistics datasets are adopted in this study, including a package delivery dataset from \citep{wu2023lade}, and a collected food delivery order datasets: (1) \textbf{Package delivery} \citep{wu2023lade}: this data is provided by the Cainiao Network in China, containing 6 months delivery orders from five cities in China;
(2) \textbf{Food delivery}: this data is provided by four different food delivery platforms in US. from 2021 to 2024. We select three representative cities with high penetration rates.

\begin{table}[!htbp]
\caption{Total number of ROIs for each city.}
\label{tab:data_summary}
\centering
\begin{small}
    \renewcommand{\multirowsetup}{\centering}
    \setlength{\tabcolsep}{1pt}
    \resizebox{0.8\textwidth}{!}{
    \begin{tabular}{l|c|c|c|c|c|c|c|c}
    \toprule
    City & Shanghai & Hangzhou & Chongqing & Jilin  & Yantai & Los Angeles & New York & San Francisco \\
    \midrule
    \# of ROIs &30 & 31& 30 &14&30 & 17& 22&19 \\
    \bottomrule
    \end{tabular}}
\end{small}
\end{table}

The entire area of each city is divided into several geospatial regions for logistics management. The datasets provide event-specific information, such as the order ID, pickup time, location, GPS points, courier information, delivery event, etc.
Each order is attached to the corresponding region of interests (ROIs) according to its pickup GPS record.
For package data, we set a time interval of 1 hour and aggregated the orders within this specified window. For food data, due to the sparsity of this data record, we select a 12-hour interval. 
In both datasets, we calculate the total number of pickups, i.e., the pickup demand.
In following sections, we adopt both package and food delivery data for model comparisons. 
{Individual orders are anonymized and aggregated into regions to eliminate privacy breaches.}
For subsequent model analysis and discussion, we mainly adopt the package delivery data for experiments.
Total number of regions for each city is summarized in Tab. \ref{tab:data_summary}.

\subsection{Experiment Settings}
\textbf{Evaluation scenarios}. To illustrate the efficacy of our methodology, we examine three scenarios: (1) end-to-end demand joint estimation and forecasting in a single city; (2) zero-shot transfer to new cities with comprehensive observations, wherein both the source and target cities possess complete historical data; and (3) zero-shot transfer to new cities with full observations, in which both the source and target cities have newly introduced regions lacking historical data.

\textbf{Baseline models}. Apart from the historical average (HA) method, we select several representative models from highly related literature in machine learning studies, including: diffusion convolutional recurrent neural network (DCRNN) \citep{DCRNN}, spatial-temporal convolutional neural network \citep{STGCN}, graph wavenet (GWNET) \citep{GWNET}, graph learning neural network (MTGNN) \citep{MTGNN}, inductive graph neural network (IGNNK) \citep{IGNNK}, spatial aggregation and temporal convolutional network (SATCN) \citep{SATCN}, message-passing gated recurrent unit (MPGRU) \citep{MPGRU}, and graph recurrent imputation network (GRIN) \citep{GRIN}. Particularly, DCRNN, STGCN, GWNET, MTGNN are originally designed for forecasting tasks; While IGNNK, SATCN, MPGRU, and GRIN are designed for kriging and imputation problems. We evaluate both of the two categories for a fair comparison.

\textbf{Tasks setups}.
Models are trained to predict the future demand of all locations in the next 24 steps, given a 24-step historical window of existing regions. For Shanghai, Hangzhou, Chongqing and Yantai, we randomly choose 10 regions as newly developed regions for test. For Jilin, Los Angeles, New York and San Francisco, the number of new regions is set to 5.
Note that the unknown testing regions are unavailable for all models during training and only accessible for testing purposes.
Following \cite{wu2023lade}, we use the 6:2:2 ratio for training, evaluation and test sets based on the chronological order of the timestamps. To assess the performance of the above models, we report the metrics of Mean Absolute Error (MAE) and Root Mean Squared Error (RMSE).

\textbf{Hyperparameters}.
For LLMs, we mainly employ an open-source model called Llama \citep{touvron2023Llama}. Another model called BERT \citep{devlin2018bert} is also used for comparison.
For \textsc{Impel}, the LLM encoding dimension is 4096 according to the default setting of Llama. The node embedding dimension is set to 32. We totally adopt 3 layers of feedforward layers and 1 spatiotemporal message-passing layer. The hidden dimension of the sequential encoder is 64 for all data.
The number of masked regions during training is set to 6 for Shanghai, Hangzhou, Chongqing, and Yantai, and 3 for others. As for hyperparameters of other baselines, we tune their parameters according to the suggested values from their original papers.

\textbf{Platforms}. All experiments are implemented on a single NVIDIA RTX A6000 GPU using PyTorch.

\begin{table}[htbp]
\caption{Experimental results of spatiotemporal delivery demand joint estimation and prediction in package delivery datasets.}
\label{tab:result_main_lade}
\centering
\begin{small}
    \renewcommand{\multirowsetup}{\centering}
    \setlength{\tabcolsep}{1pt}
    \resizebox{0.8\textwidth}{!}{
    \begin{tabular}{l|cc|cc|cc|cc|cc}
    \toprule
     \multirow{2}{*}{Models} & \multicolumn{2}{c|}{Shanghai} & \multicolumn{2}{c|}{Hangzhou} & \multicolumn{2}{c|}{Chongqing} & \multicolumn{2}{c|}{Jilin}  & \multicolumn{2}{c}{Yantai} \\
    \cmidrule(lr){2-11} 
    & MAE & RMSE & MAE & RMSE & MAE & RMSE & MAE & RMSE & MAE & RMSE  \\
    \midrule
    HA  & 6.96 & 16.41 & 8.94 & 20.56 & 4.00 & 8.85 & 2.27 & 4.85 & 3.82 & 8.42\\
    DCRNN \citep{DCRNN} & 5.65 & 11.86 & 7.33 & 14.59 & 3.53 & 6.15  & 2.05 & 3.37 & 3.21 & 6.17 \\
    STGCN \citep{STGCN} & 5.07 & 11.62 & 6.38 & 14.26 & 2.99 & 6.00  & 1.53 & \cellcolor{gray!20}{\underline{2.84}} & 2.80 & 5.79 \\
    GWNET \citep{GWNET} & 5.22 & 11.67 & 7.99 & 15.90 & 3.06 & 6.03  & 1.64 & 3.06 & 2.93 & 6.01\\
    MTGNN \citep{MTGNN} & 5.09 & 11.56 & \cellcolor{gray!20}{\underline{6.23}} & \cellcolor{gray!20}{\underline{13.89}} & \cellcolor{gray!20}{\underline{2.97}} & 5.91 & \cellcolor{gray!20}{\underline{1.52}} & \cellcolor{gray!20}{\underline{2.84}} & \cellcolor{gray!20}{\underline{2.73}} & 5.70 \\
    IGNNK \citep{IGNNK} & 5.22 & 11.50 & 7.25 & 15.06 & 3.22 & 6.06  & 2.22 & 3.71 & 2.97 & 5.93 \\
    SATCN \citep{SATCN} & \cellcolor{gray!20}{\underline{4.75}} & \cellcolor{gray!20}{\underline{9.38}}  & 7.64 & 14.77  & 3.04 & \cellcolor{gray!20}{\underline{5.27}} & 1.58 & \cellcolor{gray!20}{\underline{2.84}} & 2.83 & \cellcolor{gray!20}{\underline{5.02}} \\
    MPGRU \citep{MPGRU} & 6.30 & 13.43 & 7.95& 16.03 & 3.91 & 7.60 & 1.94 & 3.61 & 3.58 &7.45 \\
    GRIN \citep{GRIN}& 5.08 & 11.64 & 6.30 & 14.56 & 3.05  & 6.08 & 1.54 & 2.90 & 2.86 &6.02 \\
    \midrule
    \textsc{Impel} (Ours) & \cellcolor{orange!20}{\textbf{3.76}} & \cellcolor{orange!20}{\textbf{7.93}}  & \cellcolor{orange!20}{\textbf{4.52}} & \cellcolor{orange!20}{\textbf{9.90}} & \cellcolor{orange!20}{\textbf{2.47}} & \cellcolor{orange!20}{\textbf{4.92}} & \cellcolor{orange!20}{\textbf{1.39}} & \cellcolor{orange!20}{\textbf{2.51}} & \cellcolor{orange!20}{\textbf{2.23}} & \cellcolor{orange!20}{\textbf{4.18}} \\
    Improvement ($\%$) & \textcolor{black!30!green}{\textbf{20.8}} & \textcolor{black!30!green}{\textbf{15.4}} & \textcolor{black!30!green}{\textbf{27.4}} & \textcolor{black!30!green}{\textbf{28.7}} & \textcolor{black!30!green}{\textbf{16.8}} & \textcolor{black!30!green}{\textbf{6.64}} & \textcolor{black!30!green}{\textbf{8.55}} &\textcolor{black!30!green}{\textbf{11.6}}  & \textcolor{black!30!green}{\textbf{18.3}} & \textcolor{black!30!green}{\textbf{16.7}} \\
    \bottomrule
    \end{tabular}}
\end{small}
\end{table}

\subsection{Results on City-wide Demand Joint Estimation and Prediction}

\subsubsection{Task 1: Results of Model Comparison in Single Cities.}

The results of model comparison in the two datasets are shown in Tabs. \ref{tab:result_main_lade} and \ref{tab:result_main_food}. As can be seen, our model consistently outperforms baselines by a large margin, especially in mega cities like Shanghai, Hangzhou, and Los Angeles. Since the joint estimation and prediction task is more challenging than the conventional time series forecasting problem, some strong baselines from the literature (such as DCRNN and GWNET) are less competitive. For models that are originally designed for imputation or kriging, i.e., GRIN and IGNNK, they are surpassed by both our model and other forecasting-based models. Surprisingly, even without complex sequential processing modules, such as RNNs and TCNs, \textsc{Impel} still achieves the best performance in all cases.

\begin{table}[htbp]
\caption{Experimental results of spatiotemporal delivery demand joint estimation and prediction in food delivery datasets.}
\label{tab:result_main_food}
\centering
\begin{small}
    \renewcommand{\multirowsetup}{\centering}
    \setlength{\tabcolsep}{8pt}
    \resizebox{0.8\textwidth}{!}{
    \begin{tabular}{l|cc|cc|cc}
    \toprule
     \multirow{2}{*}{Models} & \multicolumn{2}{c|}{Los Angeles} & \multicolumn{2}{c|}{New York} & \multicolumn{2}{c}{San Francisco}  \\
    \cmidrule(lr){2-7} 
    & MAE & RMSE & MAE & RMSE & MAE & RMSE  \\
    \midrule
    HA  & 0.602 & 1.157 & 0.792 & 1.400 & 0.438 & 0.766 \\
    DCRNN \citep{DCRNN} & 0.481 & 0.983 & 0.602 & 1.132 & 0.328 & 0.644 \\
    STGCN \citep{STGCN} & 0.466 & 0.903 & \cellcolor{gray!20}{\underline{0.569}} & \cellcolor{gray!20}{\underline{1.050}} & 0.297 & 0.623 \\
    GWNET \citep{GWNET} & 0.457 & 0.900 & 0.614 & 1.170 & 0.298 & 0.620  \\
    MTGNN \citep{MTGNN} & \cellcolor{gray!20}{\underline{0.441}} & \cellcolor{gray!20}{\underline{0.896}} & 0.589 & 1.093 & \cellcolor{gray!20}{\underline{0.295}} & 0.623 \\
    IGNNK \citep{IGNNK} &0.448 & 0.907 & 0.573 & 1.062 & 0.301 & 0.616 \\
    SATCN \citep{SATCN} &0.471 &  0.934& 0.594 & 1.061 & 0.328 & \cellcolor{gray!20}{\underline{0.601}} \\
    MPGRU \citep{MPGRU} & 0.458 &  0.920 & 0.589 & 1.104 &  0.299 & 0.625 \\
    GRIN \citep{GRIN}   & 0.444 & 0.906 & 0.579 &  1.075 & 0.304 & 0.625 \\
    \midrule
    \textsc{Impel} (Ours) & \cellcolor{orange!20}{\textbf{0.321}} & \cellcolor{orange!20}{\textbf{0.780}}  & \cellcolor{orange!20}{\textbf{0.507}} & \cellcolor{orange!20}{\textbf{0.907}} & \cellcolor{orange!20}{\textbf{0.280}} & \cellcolor{orange!20}{\textbf{0.576}}  \\
    Improvement ($\%$) & \textcolor{black!30!green}{\textbf{27.2}} & \textcolor{black!30!green}{\textbf{12.9}} & \textcolor{black!30!green}{\textbf{10.9}} & \textcolor{black!30!green}{\textbf{13.6}} & \textcolor{black!30!green}{\textbf{5.08}} & \textcolor{black!30!green}{\textbf{4.16}}  \\
    \bottomrule
    \end{tabular}}
\end{small}
\end{table}

\begin{figure}[!htbp]
\centering
\begin{subfigure}[b]{0.48\textwidth}
\centering
\includegraphics[scale=0.38]{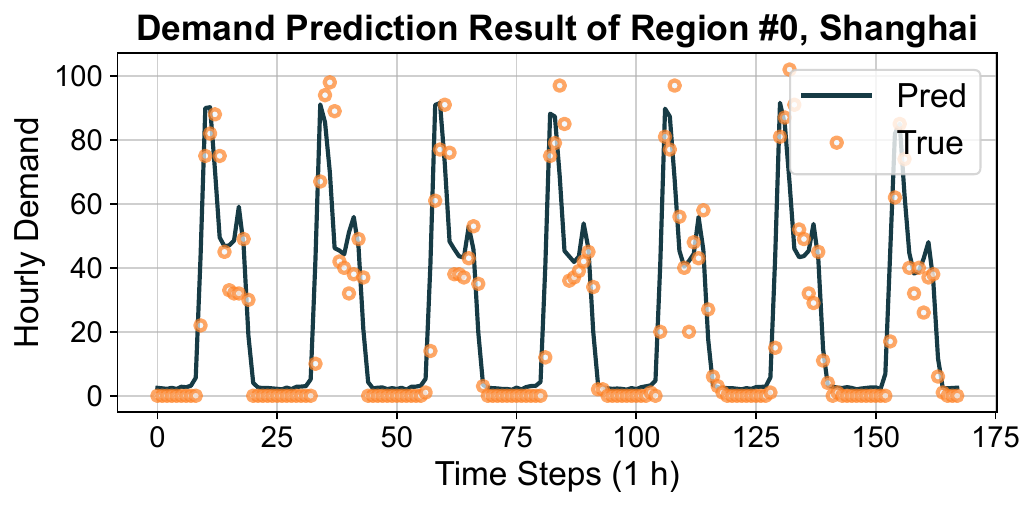}
\captionsetup{skip=1pt}
\caption{Region \#0, Shanghai}
\end{subfigure}
\hfill
\begin{subfigure}[b]{0.48\textwidth}
\centering
\includegraphics[scale=0.38]{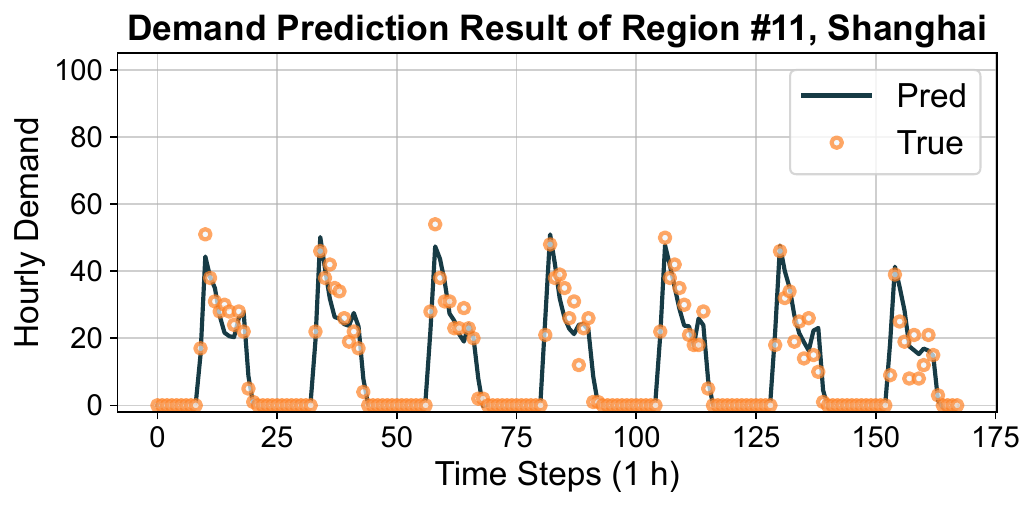}
\captionsetup{skip=1pt}
\caption{Region \#11, Shanghai}
\end{subfigure}
\hfill
\begin{subfigure}[b]{0.48\textwidth}
\centering
\includegraphics[scale=0.38]{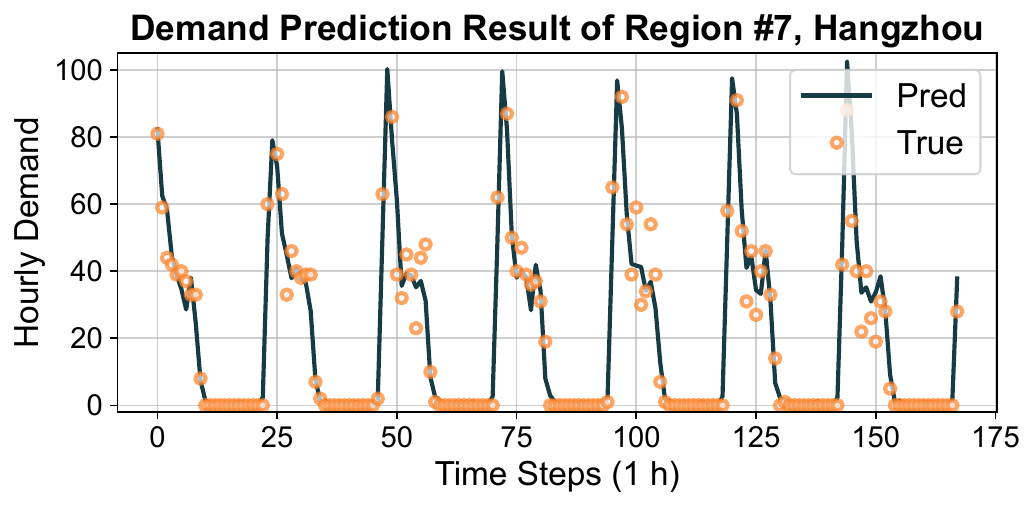}
\captionsetup{skip=1pt}
\caption{Region \#7, Hangzhou}
\end{subfigure}
\hfill
\begin{subfigure}[b]{0.48\textwidth}
\centering
\includegraphics[scale=0.38]{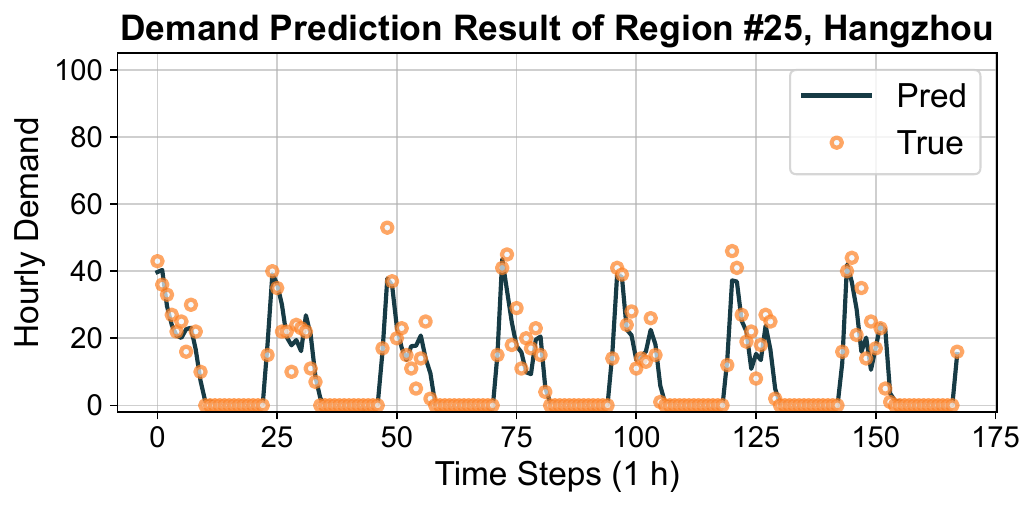}
\captionsetup{skip=1pt}
\caption{Region \#25, Hangzhou}
\end{subfigure}
\captionsetup{skip=1pt}
\caption{Visualization examples of demand estimation (prediction) in different regions.}
\label{fig:example}
\end{figure}

Fig. \ref{fig:example} provides several visualization examples that are randomly chosen from the prediction results of our model. Different regions and cities show distinct demand patterns, but share some common structures such as periodicity. Furthermore, the demand time series is sparsely sampled, with a large number of time points absent. Some data imputation methods that can be integrated in deep learning architectures can be adopted to tackle the sparse input sequence \citep{nie2024imputeformer}.

\subsubsection{Task 2: Model Transfers to New Cities with Full Observations}


In addition to the accuracy of estimation and forecasting, transferability is another prominent property of the proposed model. In this section, we evaluate two aspects of model-based transfer strategy. First, we train a model on the full region sets in a source city, then we employ it to a new city with full observations, i.e., without new regions. This model needs to outputs the prediction results directly, without tuning any parameter. Second, we train and transfer models in source and target cities that both have new regions without observations.

\begin{table}[htbp]
\caption{Experimental results of model transfer in new cities with full measurement (zero-shot prediction). We remove the spatial message passing layers in \textsc{Impel} in this case.}
\label{tab:result_transfer_full}
\centering
\begin{small}
    \renewcommand{\multirowsetup}{\centering}
    \setlength{\tabcolsep}{1pt}
    \resizebox{0.8\textwidth}{!}{
    \begin{tabular}{l|cc|cc|cc|cc|cc}
    \toprule
     Models & \multicolumn{2}{c|}{\textsc{Impel} (Ours)} & \multicolumn{2}{c|}{MTGNN} & \multicolumn{2}{c|}{IGNNK} & \multicolumn{2}{c|}{STGCN}  & \multicolumn{2}{c}{GRIN} \\
    \cmidrule(lr){2-11} 
    Source $\mapsto$ Target& MAE & RMSE & MAE & RMSE & MAE & RMSE & MAE & RMSE & MAE & RMSE  \\
    \midrule
    Shanghai $\mapsto$ Hangzhou & \cellcolor{orange!20}{\textbf{3.35}} & \cellcolor{orange!20}{\textbf{6.42}} &  4.07 & \cellcolor{gray!20}{\underline{7.32}} & 5.64 & 9.00 & 5.55 &  9.31 & \cellcolor{gray!20}{\underline{3.75}} &7.33 \\
    Shanghai $\mapsto$ Chongqing & \cellcolor{orange!20}{\textbf{2.26}} & \cellcolor{orange!20}{\textbf{4.44}} & 2.62 & 4.89 & 3.25 & 5.22  & 2.91 &  4.91 & \cellcolor{gray!20}{\underline{2.51}} & \cellcolor{gray!20}{\underline{4.86}} \\
    Shanghai $\mapsto$ Yantai & \cellcolor{orange!20}{\textbf{2.21}} & \cellcolor{orange!20}{\textbf{4.14}} & \cellcolor{gray!20}{\underline{2.52}} & \cellcolor{gray!20}{\underline{4.68}} & 3.47 & 5.66  & 2.79 & 4.77 & 2.56 & 4.87 \\
    Hangzhou $\mapsto$ Shanghai & \cellcolor{orange!20}{\textbf{2.90}}  & \cellcolor{orange!20}{\textbf{6.20}} & 3.12 & 6.38 & \cellcolor{gray!20}{\underline{3.04}} & \cellcolor{gray!20}{\underline{6.26}} & 3.09 & 6.54 & 3.11 & 6.53 \\
    Hangzhou $\mapsto$ Chongqing & \cellcolor{orange!20}{\textbf{2.10}} & \cellcolor{orange!20}{\textbf{4.25}} & 2.21 & 4.37 & 2.21 & \cellcolor{gray!20}{\underline{4.32}}  & \cellcolor{gray!20}{\underline{2.17}} &  4.37 & 2.20 & 4.47 \\
    Hangzhou $\mapsto$ Yantai & \cellcolor{orange!20}{\textbf{2.13}} & \cellcolor{orange!20}{\textbf{4.08}} & 2.26 & 4.33 & 2.23  & \cellcolor{gray!20}{\underline{4.17}} &  \cellcolor{gray!20}{\underline{2.21}} &4.28 & 2.34 & 4.69 \\
    \bottomrule
    \end{tabular}}
\end{small}
\end{table}

Tab. \ref{tab:result_transfer_full} shows the results of model transfers in the entire dataset (both the source and the target sets). This zero-shot prediction task is challenging, as different cities can have distinct demand patterns in both space and time. Through this experiment, we would like to examine whether the model learns some shared patterns of demand distribution. As demonstrated, our model shows superior transferability than other models in this pure forecasting task. Results of different source-target city pairs also indicate that different cities exhibit various demand patterns, but our model can capture the shared structures and learn to narrow this gap.

\subsubsection{Task 3: Model Transfers to New Cities with New {Regions}}
\begin{table}[htbp]
\caption{Experimental results of model transfer in new cities with partial measurement (both the source city and the target city have unobserved regions).}
\label{tab:result_transfer_partial}
\centering
\begin{small}
    \renewcommand{\multirowsetup}{\centering}
    \setlength{\tabcolsep}{1pt}
    \resizebox{0.8\textwidth}{!}{
    \begin{tabular}{l|c|cc|cc|cc|cc}
    \toprule
    \multirow{2}{*}{Source $\mapsto$ Target}  & Models & \multicolumn{2}{c|}{\textsc{Impel} (Ours)} & \multicolumn{2}{c|}{MTGNN} & \multicolumn{2}{c|}{IGNNK} & \multicolumn{2}{c}{STGCN}  \\
    \cmidrule(lr){2-10} 
    & \# of new regions & MAE & RMSE & MAE & RMSE & MAE & RMSE & MAE & RMSE  \\
    \midrule
    \multirow{2}{*}{Shanghai $\mapsto$ Hangzhou} & 5 & \cellcolor{orange!20}{\textbf{5.33}} & \cellcolor{orange!20}{\textbf{9.26}} & \cellcolor{gray!20}{\underline{6.51}}  & \cellcolor{gray!20}{\underline{13.70}} & 6.53 & 13.87 & 7.39 & 14.73 \\
     & 10 & \cellcolor{orange!20}{\textbf{5.54}} & \cellcolor{orange!20}{\textbf{9.50}} & 7.14 & \cellcolor{gray!20}{\underline{14.29}} & \cellcolor{gray!20}{\underline{7.13}} & 14.52 & 8.02 & 15.44 \\
    \multirow{2}{*}{Shanghai $\mapsto$ Chongqing} &5  &  \cellcolor{orange!20}{\textbf{2.92}} & \cellcolor{orange!20}{\textbf{4.79}} &  3.14 &  5.73 & \cellcolor{gray!20}{\underline{3.00}} & \cellcolor{gray!20}{\underline{5.52}} & 3.33 & 6.00 \\
     &10 & \cellcolor{orange!20}{\textbf{3.09}}  & \cellcolor{orange!20}{\textbf{5.02}} & 3.45 & 6.16 & \cellcolor{gray!20}{\underline{3.35}} & \cellcolor{gray!20}{\underline{5.98}} & 3.63 & 6.43 \\
    \multirow{2}{*}{Shanghai $\mapsto$ Yantai} & 5& \cellcolor{orange!20}{\textbf{2.72}} & \cellcolor{orange!20}{\textbf{4.47}} &  2.96 & 5.29 & \cellcolor{gray!20}{\underline{2.93}} & \cellcolor{gray!20}{\underline{5.19}}  & 3.25 & 5.90 \\
    & 10& \cellcolor{orange!20}{\textbf{2.82}} & \cellcolor{orange!20}{\textbf{4.57}} &  \cellcolor{gray!20}{\underline{3.38}} & 6.26 & \cellcolor{gray!20}{\underline{3.38}}  & \cellcolor{gray!20}{\underline{6.19}} & 3.68 & 6.80 \\
    \bottomrule
    \end{tabular}}
\end{small}
\end{table}

When there exist regions without historical data in both source and target domains, the challenge of model transfer is further intensified. Tab. \ref{tab:result_transfer_partial} examines the performance of a mode trained in Shanghai and transferred to new cities with different numbers of new regions. As expected, the prediction error is larger than the results in Tab. \ref{tab:result_transfer_full} due to the existence of unobserved regions in both the training set and the test set. In this scenario, our model shows more significant superiority over baselines. This can be ascribed to the generality of LLM-based location encoding that has comprehensive and fundamental geospatial knowledge shared across cities.

\section{Discussions}\label{sec:discussion}
This section provides comprehensive discussions to enhance the interpretability of the proposed framework. We first discuss and interpret the role of LLM-based encodings. Then the impact of modular designs and the choice of hyperparameters of analyzed. Finally, the rationality of the joint estimation and prediction is verified.


{\subsection{Benchmarking the LLM-based Geolocation Encoding}
To justify the significance of LLM-based geolocation encoding in delivery modeling, we perform experiments to compare it with other geo-encoding methods. 
This section contains the following evaluations:
\begin{enumerate}
    \item Comparison with state-of-the-art geographic embedding methods, including Space2Vec \citep{Space2Vec} and Sphere2Vec \citep{Sphere2vec};
    \item Encoding POIs with large embedding models, such as the General Text Embeddings (GTE) \citep{GTE};
    \item Using one-hot encoding to process POI features;
    \item Only encoding of target address with LLMs.
\end{enumerate}

\begin{table}[htbp]
\caption{{Evaluation results of different geo-encoding methods.}}
\label{tab:geo-encoding}
\centering
\begin{small}
    \renewcommand{\multirowsetup}{\centering}
    \setlength{\tabcolsep}{1pt}
    \resizebox{0.8\textwidth}{!}{
    \begin{tabular}{l|cc|cc|cc|cc|cc}
    \toprule
     \multirow{2}{*}{Models} & \multicolumn{2}{c|}{Shanghai} & \multicolumn{2}{c|}{Hangzhou} & \multicolumn{2}{c|}{Chongqing} & \multicolumn{2}{c|}{Jilin}  & \multicolumn{2}{c}{Yantai} \\
    \cmidrule(lr){2-11} 
    & MAE & RMSE & MAE & RMSE & MAE & RMSE & MAE & RMSE & MAE & RMSE  \\
    \midrule
    \textsc{Impel} (Ours) & \cellcolor{orange!20}{\textbf{3.76}} & \cellcolor{orange!20}{\textbf{7.93}}  & \cellcolor{orange!20}{\textbf{4.52}} & \cellcolor{orange!20}{\textbf{9.90}} & \cellcolor{orange!20}{\textbf{2.47}} & \cellcolor{orange!20}{\textbf{4.92}} & \cellcolor{orange!20}{\textbf{1.39}} & \cellcolor{orange!20}{\textbf{2.51}} & \cellcolor{orange!20}{\textbf{2.23}} & \cellcolor{orange!20}{\textbf{4.18}} \\
    \midrule
    Space2Vec  & 4.28 & 7.93 &5.93  & 10.75 &6.47  & 10.28 & 1.52 & 2.68 & 3.41 & 6.61  \\
    Sphere2Vec & 4.33 & 8.23 & 10.26 & 20.17 & 3.13 & 5.54 & 1.47 & 2.67  & 3.55 & 6.73 \\
    LLM+Address & 5.95 & 14.17 & 4.99 & 10.48 & 2.96 & 5.38 & 1.98 & 4.08 & 2.94 & 5.22 \\
    POI Only & 6.98 & 16.00 & 11.44 & 22.07 & 7.67 & 12.03 & 2.69 & 5.48 &4.06  & 7.79 \\
    GTE+POI & 4.08 & 8.71 & 5.32 & 11.88 & 3.04 & 5.82 & 1.90 & 4.05 & 2.89 & 4.86 \\
    \bottomrule
    \end{tabular}}
\end{small}
\end{table}

Results are shown in Table \ref{tab:geo-encoding}. As can be seen, our model consistently achieves the best performances. Pretrained geographic methods such as Space2Vec cannot provide effective contextual information for delivery modeling. Additionally, encoding only the coordinates (address) or using other methods to process POIs cannot achieve desirable performances.

\subsection{The generality of LLM-based Geolocation Encoding in Enhancing Demand Prediction}

Recall that our model has a simple-yet-effective architecture: it bases on stacked spatiotemporal message passing layers that are easy to compute. This efficiency is achieved by the effectiveness of LLM-based encoding.
To evaluate whether the enhancement of LLM-based encoding is generic, we equip other baselines with the same encoding and report their performances. 
{The results presented in Tab. \ref{tab:result_w/_llm} indicate that all baseline architectures can benefit from the embedding, with improvements exceeding 10\%. This result means that better prediction accuracy can be achieved by adopting more advanced backbone architectures.}

\begin{table}[htbp]
\caption{{Performance gains by equipping with LLM-based encoding and graph.}}
\label{tab:result_w/_llm}
\centering
\begin{small}
    \renewcommand{\arraystretch}{0.5} 
    \setlength{\tabcolsep}{8pt}
    \resizebox{0.7\columnwidth}{!}{
    \begin{tabular}{l|cc|cc|c}
    \toprule
     \multirow{2}{*}{Models} & \multicolumn{2}{c|}{Shanghai} & \multicolumn{2}{c|}{Hangzhou} & \multicolumn{1}{c}{IMP}\\
    \cmidrule(lr){2-6} 
    & MAE & RMSE & MAE & RMSE & (\%)\\
    \midrule
    DCRNN  & 5.65 & 11.86 & 7.33 & 14.59   & \multirow{2}{*}{14.83\%}\\
    \cellcolor{gray!20}{+LLM Enc.}  & \textbf{4.86} & \textbf{9.98} & \textbf{6.34}  & \textbf{12.55} & \\
    \midrule
    STGCN  & 5.07 & 11.62 & 6.38 & 14.26 & \multirow{2}{*}{18.64\%}\\
    \cellcolor{gray!20}{+LLM Enc.}  & \textbf{4.47} & \textbf{9.34} & \textbf{5.36}  & \textbf{10.39} & \\
    \midrule
    GWNET  & 5.22 & 11.67 & 7.99 & 15.90  & \multirow{2}{*}{21.04\%}\\
    \cellcolor{gray!20}{+LLM Enc.}  & \textbf{4.58} & \textbf{9.25} & \textbf{6.03}  & \textbf{11.69} & \\
    \midrule
    MTGNN  & 5.09 & 11.56 & 6.23 & 13.89  & \multirow{2}{*}{21.92\%}\\
    \cellcolor{gray!20}{+LLM Enc.}  & \textbf{4.05} & \textbf{8.15} & \textbf{5.45}  & \textbf{10.38} & \\
    \midrule
    IGNNK  & 5.22 & 11.50 & 7.25 & 15.06  & \multirow{2}{*}{25.52\%}\\
    \cellcolor{gray!20}{+LLM Enc.} & \textbf{3.95} & \textbf{7.90} & \textbf{5.88}  & \textbf{11.21} & \\
    \midrule    
    SATCN  & 4.75 & 9.38 & 7.64 & 14.77  & \multirow{2}{*}{13.68\%}\\
    \cellcolor{gray!20}{+LLM Enc.}  & \textbf{4.58} & \textbf{8.85} & \textbf{6.12}  & \textbf{11.00} & \\
    \bottomrule
    \end{tabular}}
\end{small}
\end{table}
}

\subsection{Understanding the Encoding from LLMs}

According to the discussions mentioned above, the role of LLM-based encoding is pivotal for our model.
Therefore, we conduct an analysis to compare different LLM backbones and to determine whether to use the reinforcement learning from human feedback (RLHF) scheme \citep{bai2022training}. As an additional comparison, we also construct a Laplacian-based encoding that adopts the random-walk positional encoding for GNNs \citep{dwivedi2021graph}, and a random encoding sampled from a Gaussian distribution \citep{abboud2020surprising}, which are widely used in graph machine learning studies.

\begin{figure}[!htbp]
  \centering
  \captionsetup{skip=1pt}
  \includegraphics[width=0.9\columnwidth]{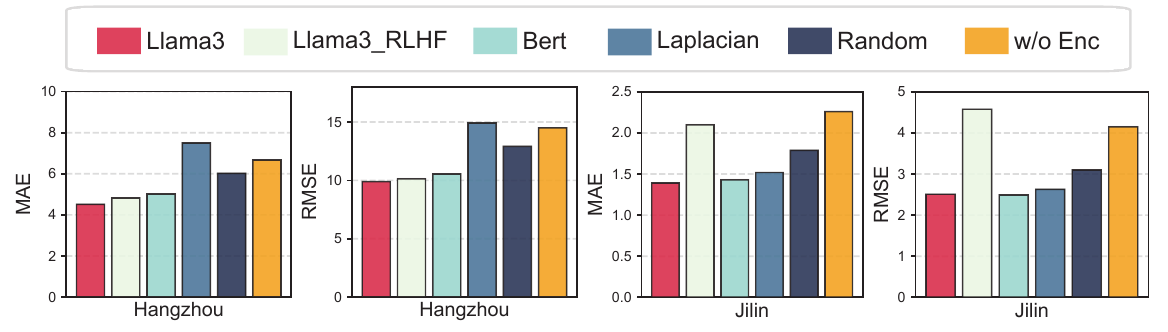}
  \caption{Ablation study on the choice of region encoding.}
  \label{fig:ablation_llm}
\end{figure}

There are three observations in Fig. \ref{fig:ablation_llm}. (1) Llama and BERT achieve comparable performances, showing lower error than graph-based encodings (i.e., ``Laplacian'' and ``Random''). (2) After being processed with RLHF, the \textit{performance degrades a lot}, which is counter-intuitive. We claim that the embedding from the output layer of LLMs tends to align with the node embedding of GNNs in the feature space. The intervention of RLHF can distort this natural alignment to some extent. (3) After removing the encoding, our model cannot achieve desirable results, which echos our collective-individual designs in section \ref{subsec:llm_integration}.


\begin{figure}[!htbp]
  \centering
  \captionsetup{skip=1pt}
  \includegraphics[width=0.66\columnwidth]{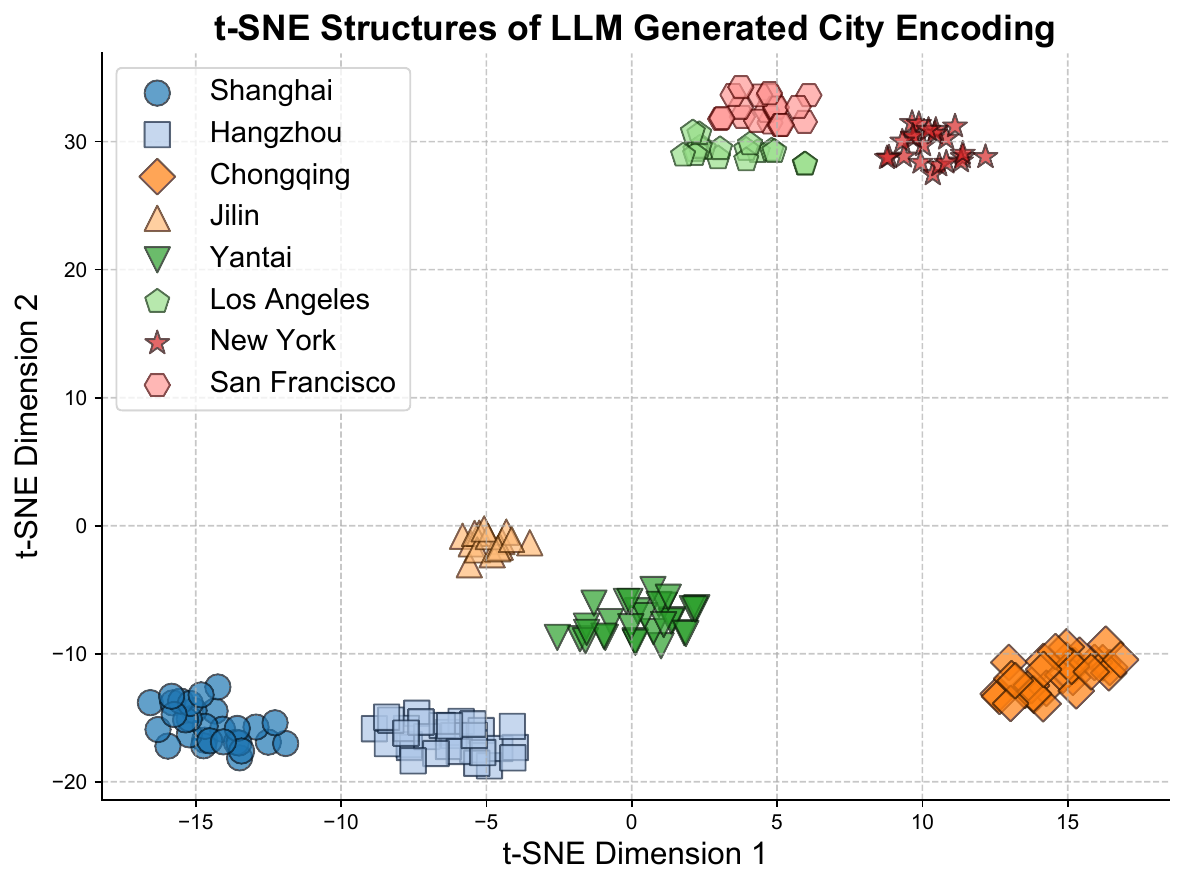}
  \caption{The t-SNE structures of LLM encoding of eight cities.}
  \label{fig:llmvec_viz}
\end{figure}

An intriguing research question pertains to the manner in which LLMs comprehend the geospatial relationship. To initiate an investigation into this matter, we present in Fig. \ref{fig:llmvec_viz} a visualization of the encoding produced by LLMs, employing the t-SNE dimensional reduction technique. Interestingly, the encoding can differentiate different cities, and regions belonging to the same city are aggregated closely. Cities in different countries are relatively far apart, which is in line with human understanding.

\subsection{Model analysis}

\subsubsection{Ablation Study}
To justify the rationality of the integrated framework, we perform ablation study to examine the following model variations by removing or replacing existing components in \textsc{Impel}:
\begin{enumerate}
    \item w/o LLM-graph: we remove the LLM-based graph construction and message passing layers;
    \item w adjacency graph: we replace the LLM-based graph with a proximity {(distance)}-based graph;
    \item w/o updating layers: we remove the dense feedforward layers after the message passing;
    \item First update then propagate: we implement feedforward first, then message passing.
\end{enumerate}

\begin{figure}[!htbp]
  \centering
  \captionsetup{skip=1pt}
  \includegraphics[width=0.8\columnwidth]{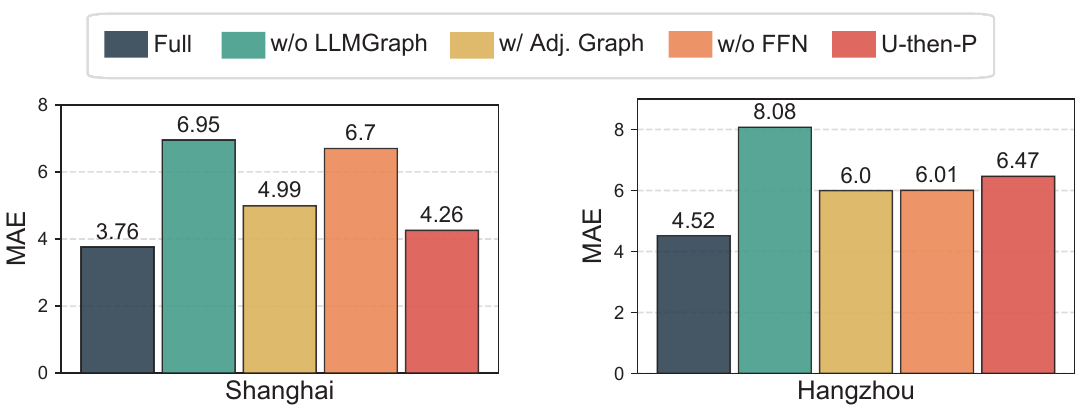}
  \caption{Ablation study on the choice of modules.}
  \label{fig:ablation_arch}
\end{figure}

Fig. \ref{fig:ablation_arch} displays the results of architectural ablation studies. It is apparent that the LLM-based functional graph plays a pivotal role in model performance. On the contrary, the proximity-based graph is less informative in this task. Additionally, the feedforward layers are also indispensable and can make more contributions when equipped after the message-passing layers.

\subsubsection{Sensitivity Analysis}
\begin{figure}[!htbp]
  \centering
  \captionsetup{skip=1pt}
  \includegraphics[width=0.8\columnwidth]{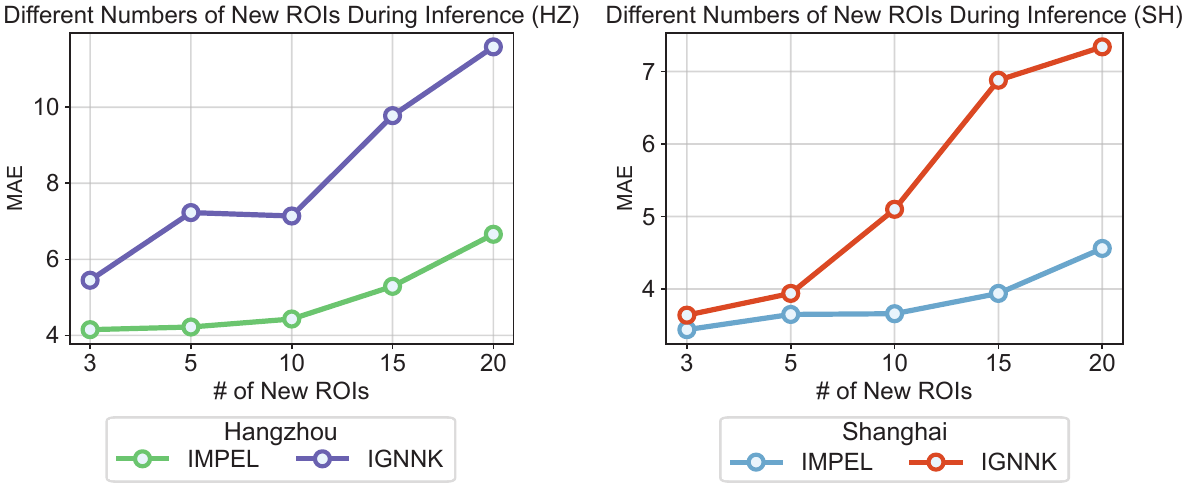}
  \caption{Model performance under different numbers of newly added regions.}
  \label{fig:different_n_u}
\end{figure}

The plot in Fig. \ref{fig:different_n_u} shows the influence of different observation rates, in other words, different numbers of new regions. As expected, the task becomes more challenging as more regions are developed. But our model is more robust to missing observations than the baseline IGNNK in these scenarios. 

\begin{figure}[!htbp]
  \centering
  \captionsetup{skip=1pt}
  \includegraphics[width=0.8\columnwidth]{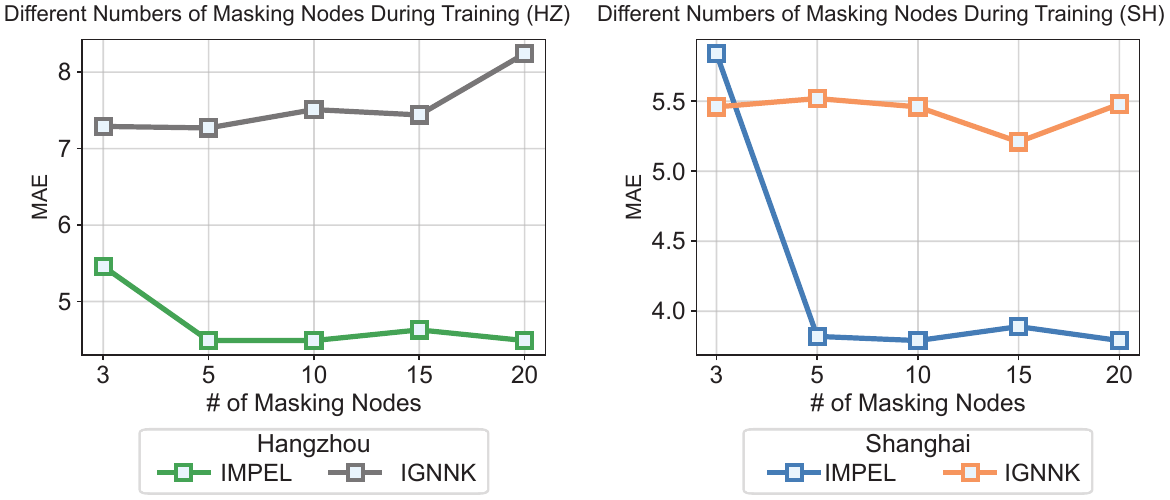}
  \caption{Model performance under different numbers of masking nodes during training.}
  \label{fig:different_n_m}
\end{figure}

In Fig. \ref{fig:different_n_m} we further examine the impacts of the masking strategy. Both IGNNK and \textsc{Impel} share the inductive training strategy. However, \textsc{Impel} can benefit more from the increase in masking rates during training. Even with a large proportion of regions being masked, our model can still reconstruct the target set with guaranteed accuracy, which is attributed to the versatility of LLM-encoding.

\subsubsection{Comparison with Separate Estimation and Prediction.}

Furthermore, it would be beneficial to ascertain whether \textsc{Impel} enhances the kriging or forecasting capabilities. To determine this, we first train a forecasting model based on partial regions with full historical records. Then we incorporate the newly added ROIs and compare the following strategies: (1) End-to-end estimation and forecasting; (2) Direct forecasting for all regions based on the incomplete data; (3) First estimating the unknown data with a pretrained estimator, i.e., a kriging model, and then forecasting the future states of all regions. IGNNK is adopted as the pretrained kriging model; (4) First forecasting the future values of observed data, and then estimating the unobserved regions with a pretrained kriging model.

\begin{table}[htbp]
\caption{Experimental results of model in separate estimation and forecasting tasks.}
\label{tab:result_separate_tasks}
\centering
\begin{small}
    \renewcommand{\multirowsetup}{\centering}
    \setlength{\tabcolsep}{5pt}
    \resizebox{0.8\textwidth}{!}{
    \begin{tabular}{l|c|cc|cc|cc}
    \toprule
    \multirow{2}{*}{Tasks}  & \multirow{2}{*}{Models} & \multicolumn{2}{c|}{Shanghai} & \multicolumn{2}{c|}{Hangzhou} & \multicolumn{2}{c}{Chongqing}   \\
    \cmidrule(lr){3-8} 
    &  & MAE & RMSE & MAE & RMSE & MAE & RMSE  \\
    \midrule
    \multirow{5}{*}{Forecasting} & STGCN & \cellcolor{gray!20}{\underline{3.04}} & \cellcolor{gray!20}{\underline{6.42}}  & \cellcolor{gray!20}{\underline{3.01}}  & \cellcolor{gray!20}{\underline{5.98}}  & \cellcolor{gray!20}{\underline{2.16}}  & \cellcolor{gray!20}{\underline{4.38}} \\
     & DCRNN & 3.69 &  7.08 &  4.14 & 7.35  &  2.75 & 5.11 \\
     & GWNET & 3.16 & 6.56  & 3.22  & 6.32  & 2.22  &  4.45 \\
     \cmidrule(lr){2-8} 
     & \textsc{Impel} (Ours) & \cellcolor{orange!20}{\textbf{2.83}} & \cellcolor{orange!20}{\textbf{6.00}}  & \cellcolor{orange!20}{\textbf{2.76}}  &  \cellcolor{orange!20}{\textbf{5.55}} & \cellcolor{orange!20}{\textbf{2.08}}  & \cellcolor{orange!20}{\textbf{4.19}} \\
     & Improvement $(\%)$ & \textcolor{black!30!green}{\textbf{6.91}} & \textcolor{black!30!green}{\textbf{6.54}}  & \textcolor{black!30!green}{\textbf{8.31}}  & \textcolor{black!30!green}{\textbf{7.19}}  & \textcolor{black!30!green}{\textbf{3.70}} & \textcolor{black!30!green}{\textbf{4.34}}  \\
     \midrule
    \multirow{4}{*}{Estimation} & IGNNK & \cellcolor{gray!20}{\underline{3.48}}  & 10.57 &  \cellcolor{gray!20}{\underline{5.70}} &  \cellcolor{gray!20}{\underline{14.93}} & \cellcolor{gray!20}{\underline{1.91}} & 4.64  \\
     &SATCN & 3.96  & \cellcolor{gray!20}{\underline{8.12}}  & 7.81  & 15.15 &2.59 & \cellcolor{gray!20}{\underline{4.47}} \\
     \cmidrule(lr){2-8} 
     & \textsc{Impel} (Ours) & \cellcolor{orange!20}{\textbf{1.79}} & \cellcolor{orange!20}{\textbf{5.50}}  & \cellcolor{orange!20}{\textbf{2.59}}  &  \cellcolor{orange!20}{\textbf{8.50}} &  \cellcolor{orange!20}{\textbf{0.91}} &  \cellcolor{orange!20}{\textbf{2.76}}\\
     & Improvement $(\%)$ & \textcolor{black!30!green}{\textbf{48.6}} & \textcolor{black!30!green}{\textbf{32.3}}  &  \textcolor{black!30!green}{\textbf{54.6}} &  \textcolor{black!30!green}{\textbf{43.1}} & \textcolor{black!30!green}{\textbf{52.4}} &  \textcolor{black!30!green}{\textbf{38.3}} \\
    \bottomrule
    \end{tabular}}
\end{small}
\end{table}

As an initial evaluation, Tab. \ref{tab:result_separate_tasks} report the performance of separate estimation and forecasting tasks. These models are trained on respective tasks with separate objectives. It can be seen that \textsc{Impel} has a beneficial effect on the performance of both tasks, with a particularly notable improvement in the more difficult estimation task. Tab. \ref{tab:different_schemes} further sheds light on the necessity of the joint estimation and prediction framework. Separate treatment can achieve a lower training error by learning to fit the full distribution. However, they over-fit the training data and generate poor generalization performance in testing set where new regions are included. Instead, our end-to-end setting balances the capacity and robustness, achieving the highest accuracy in this hybrid task.

\begin{table}[htbp]
\caption{Different solution scheme and the effects of joint estimation and forecasting.}
\label{tab:different_schemes}
\centering
\begin{small}
    \renewcommand{\multirowsetup}{\centering}
    \setlength{\tabcolsep}{5.3pt}
    \resizebox{0.8\textwidth}{!}{
    \begin{tabular}{c|c|cc|c|cc}
    \toprule
    \multirow{3}{*}{Design}  & \multicolumn{3}{c|}{Shanghai} & \multicolumn{3}{c}{Hangzhou}  \\
     \cmidrule(lr){2-4} \cmidrule(lr){5-7}    
    & \multicolumn{1}{c|}{Training} & \multicolumn{2}{c|}{Testing} & \multicolumn{1}{c|}{Training} & \multicolumn{2}{c}{Testing} \\
    & MAE  & MAE & RMSE  & MAE  & MAE & RMSE \\
    \midrule
     \textbf{End-to-End}  & \cellcolor{orange!20}{3.48}  & \cellcolor{orange!20}{\textbf{3.76}} & \cellcolor{orange!20}{\textbf{7.93}} & \cellcolor{orange!20}{3.61}  & \cellcolor{orange!20}{\textbf{4.52}} & \cellcolor{orange!20}{\textbf{9.90}}  \\
     \midrule
     Direct Forecast  & 3.30 & 6.14 & 11.84 & 3.18 &  8.32 & 16.53   \\
    \midrule
    Estimate-then-Forecast  & 3.33 & 5.10 & 11.12 & \textbf{3.08} & 6.72 & 14.71 \\
    \midrule
    Forecast-then-Estimate  & \textbf{3.28} & 5.07 & 11.09 & 3.27 & 7.08 & 15.43 \\
    \bottomrule
    \end{tabular}}
\end{small}
\end{table}

\section{Conclusion and Future Work}\label{sec:conclusion}
Accurate and real-time demand estimation and forecasting provide timely information for decision making and strategic management in urban delivery logistics. To jointly achieve the goal of estimating demand for new regions and predicting future values for existing regions, this paper introduces a graph-based deep learning framework enhanced by large language models. An inductive training scheme is then developed to promote the transfer of the model to new regions and cities. Experiments on two real-world delivery datasets, including eight cities in China and the US, show that our model significantly outperforms state-of-the-art baselines. It also has better zero-shot transferability in new cities, is more robust to the addition of new regions, and can benefit a lot from the end-to-end learning paradigm. 
Moreover, it is evident that the LLM-based encoding, which is geospatially aware, plays a crucial role in characterizing region-specific demand patterns and facilitating transferability across cities based on the graph constructed on it.

{Although promising results have been demonstrated, there are several shortcomings need further attention. First, it is important to ensure the interpretability of the model within the logistics and delivery sector to maintain the trust of stakeholders. Our model is currently based on black-box architectures in which explainable AI techniques can be used to interpret it. Second, the estimation and transfer of demand in new regions are based on the observation of existing regions, which can lead to error accumulation.}

Further research could investigate the potential of integrating this framework with more sophisticated LLM techniques, such as chain-of-thoughts prompting \citep{wei2022chain} and retrieval-augmented generation \citep{lewis2020retrieval}, for other tasks in logistics management, including route planning and optimization, expected time of arrival estimation, and depot location optimization \citep{liu2023can}.  


\section*{Acknowledgments}
This research was sponsored by the National Natural Science Foundation of China (52125208), the National Natural Science Foundation of China's Fundamental Research Program for Young Students (524B2164), and grants from the Research Grants Council of the Hong Kong Special Administrative Region, China (Project No. PolyU/15206322 and PolyU/15227424).

\section{Appendix}


{\subsection{More explanations on the role of LLM-based encoding}
We provide more explanations on the role of LLM-derived embedding in the context of demand prediction and estimation.
Actually, the demand prediction problem (without missing observations) can generally be considered as a time series forecasting task in which deep neural networks are applied to predict future value based on historical data. In state-of-the-art time series forecasting models, three types of additional features can be collected and used as covariates \citep{cini2023taming}: (1) time-related features, such as time-of-day and day-of-week encodings; (2) region-specific variates, such as the region (sensor) identity information and property; (3) {cross-region variates}, such as region-wide relationships described by an adjacency graph. These features complement the historical-future data pairs, and this setting is commonly acknowledged in previous work \citep{DCRNN,GWNET,STGCN,GRIN,MTGNN,nie2024imputeformer}. 

For (1), we have performed supplementary experiments by adding timestamp features to the model in subsection \ref{sec:app:temporal}, and the results show that the additional temporal features do not benefit the performances significantly; For (2), we use the LLM-based location encoding for each region as specialized features; For (3), we construct a functional graph based on the LLM-based encoding to quantify the relationship between regions.
Therefore, the LLM-derived embedding can characterize rich information to improve prediction performance.
}

\subsection{Supplementary experiments}

\subsubsection{Experiments on the observation patterns}

It should be noted that the experiments described in section \ref{sec:experiments} are based on the random selection and masking of regions for evaluation purposes. This may not represent the real-world scenarios of service area expansion.
To mitigate this experimental bias, we perform additional experiments that remove a whole cluster of regions that are close to each other, which leads to a cluster missing pattern.
Note that the random seed is fixed for all baselines for the same city for a fair comparison. Tab. \ref{tab:result_sup_lade} provides the results on package delivery datasets. As expected, most of the baselines show worse performance due to the difficulty in model generalization. Interestingly, in some cases our model performs better than the random missing scenario and has a greater margin of improvement. This is due to the effectiveness of the LLM-based encoding for providing sufficient geolocation information about the target location.

\begin{table}[htbp]
\caption{{Experimental results of spatiotemporal delivery demand joint estimation and prediction in package delivery datasets. In these experiments, the unobserved locations are sampled not at random, i.e., a spatial cluster of regions that are near each other is removed.}} 
\label{tab:result_sup_lade}
\centering
\begin{small}
    \renewcommand{\multirowsetup}{\centering}
    \setlength{\tabcolsep}{0.6pt}
    \resizebox{0.7\textwidth}{!}{
    \begin{tabular}{l|cc|cc|cc|cc}
    \toprule
     \multirow{2}{*}{Models} & \multicolumn{2}{c|}{Shanghai} & \multicolumn{2}{c|}{Hangzhou} & \multicolumn{2}{c|}{Chongqing}  & \multicolumn{2}{c}{Yantai} \\
    \cmidrule(lr){2-9} 
    & MAE & RMSE & MAE & RMSE & MAE & RMSE & MAE & RMSE  \\
    \midrule
    DCRNN \citep{DCRNN} & 5.77 & 12.30 & 6.32 & 11.91 & 4.42 & 8.36 & 2.65 & \cellcolor{gray!20}{\underline{5.05}} \\
    STGCN \citep{STGCN} & 5.60 & 12.86 & 6.33 & 10.98 & 4.52 & 10.71 & \cellcolor{gray!20}{\underline{2.61}} & 5.68 \\
    GWNET \citep{GWNET} & 13.06 & 25.21 & 6.32 & 11.43 & 4.52 & 10.24  & 2.72 & 5.73 \\
    MTGNN \citep{MTGNN} & 5.56 & 12.50 & 6.09 & 11.37 & 4.57 & 10.90 & 2.77 & 5.72 \\
    IGNNK \citep{IGNNK} & 5.86 & 13.21 & 5.82 & 10.44 & 4.72 & 10.89  & 3.11 &  6.76 \\
    SATCN \citep{SATCN} & \cellcolor{gray!20}{\underline{5.06}} &  \cellcolor{gray!20}{\underline{9.31}}  & 6.57 & 11.15  & \cellcolor{gray!20}{\underline{3.64}} & \cellcolor{gray!20}{\underline{6.62}}  & 3.37 & 5.78 \\
    MPGRU \citep{MPGRU} & 6.64 & 14.55 & 7.24 & 13.11 & 5.35 & 11.89  & 3.51 &7.46 \\
    GRIN \citep{GRIN}& 5.55 & 12.76 & \cellcolor{gray!20}{\underline{5.46}} &  \cellcolor{gray!20}{\underline{10.36}} & 4.67  & 11.14 & 2.85 & 6.19 \\
    \midrule
    \textsc{Impel} (Ours) & \cellcolor{orange!20}{\textbf{3.45}} & \cellcolor{orange!20}{\textbf{7.00}}  & \cellcolor{orange!20}{\textbf{4.41}} & \cellcolor{orange!20}{\textbf{9.56}} & \cellcolor{orange!20}{\textbf{2.50}} & \cellcolor{orange!20}{\textbf{5.21}}  & \cellcolor{orange!20}{\textbf{2.30}} & \cellcolor{orange!20}{\textbf{4.73}} \\
    Improvement ($\%$) & \textcolor{black!30!green}{\textbf{31.82}} & \textcolor{black!30!green}{\textbf{24.81}} & \textcolor{black!30!green}{\textbf{19.23}} & \textcolor{black!30!green}{\textbf{7.72}} & \textcolor{black!30!green}{\textbf{31.32}} & \textcolor{black!30!green}{\textbf{21.30}} & \textcolor{black!30!green}{\textbf{11.88}} &\textcolor{black!30!green}{\textbf{6.34}}  \\
    \bottomrule
    \end{tabular}}
\end{small}
\end{table}

{
\subsubsection{Experiments on the Temporal Modeling}\label{sec:app:temporal}
A common practice in deep traffic prediction model is to process the sequence with temporal modules such as recurrent neural networks and timestamp features.
Therefore, we examine the effect of temporal techniques in our architecture by equipping it with temporal modules and timestamp features.
{For timestamp features, we use the time-of-day and day-of-week positional encodings used in previous work \citep{GRIN} and concatenate them with the input time series. For additional temporal modules including RNN and TCN, we replace the original MLP in Eq. \ref{eq:temporal_enc_final} with them.}
We report both the accuracy and {training} efficiency in Table \ref{tab:temporal}.
Performances of several baselines are also shown as a reference.
It is observed that the gain of temporal modeling is marginal. However, introducing complicated temporal techniques has increased the computational burden.
To this end, we choose our architecture for simplicity and better transferability.

\begin{table}[htbp]
\caption{{The impact of temporal modeling and computational efficiency ({training stage}) of different settings. {``w/'' denotes that a new module is added to the original architecture.}}}
\label{tab:temporal}
\centering
\begin{small}
    \renewcommand{\multirowsetup}{\centering}
    \renewcommand{\arraystretch}{0.8} 
    \setlength{\tabcolsep}{2pt}
    \resizebox{1\textwidth}{!}{
    \begin{tabular}{l|cc|ccc|cc|ccc}
    \toprule
     \multirow{2}{*}{Models} & \multicolumn{5}{c|}{Shanghai} & \multicolumn{5}{c}{Hangzhou} \\
    \cmidrule(lr){2-11} 
    & MAE & RMSE & Memory & Speed & Params. & MAE & RMSE & Memory &Speed & Params. \\
    \midrule
    \textsc{Impel} (Ours) & \cellcolor{orange!20}{\textbf{3.76}} & \cellcolor{orange!20}{\textbf{7.93}} & \cellcolor{orange!20}{\textbf{1.5 GB}} & \cellcolor{orange!20}{\textbf{1.1 s/batch}} & \cellcolor{orange!20}{\textbf{218.6 k}} & \cellcolor{orange!20}{\textbf{4.52}} & \cellcolor{orange!20}{\textbf{9.90}} & \cellcolor{orange!20}{\textbf{1.6 GB}} & \cellcolor{orange!20}{\textbf{1.2 s/batch}} & \cellcolor{orange!20}{\textbf{218.6 k}}\\
    DCRNN  & 5.65 & 11.86 & 2.3 GB & 929.7 s/batch & 2215.8 k & 7.33 & 14.59  & 2.5 GB & 1311.8 s/batch & 2215.8 k\\
    GWNET & 5.22 & 11.67 & 3.0 GB & 165.8 s/batch & 2434.7 k & 7.99 & 15.90 & 3.2 GB & 175.0 s/batch & 2435.8 k\\
    MTGNN & 5.09 & 11.56 & 1.7 GB & 52.8 s/batch & 2275.3 k & 6.23 & 13.89 & 1.8 GB & 36.9 s/batch & 2338.1 k \\
    \midrule
    w/ RNN  & 4.38 & 8.87 & 1.8 GB & 8.8 s/batch & 483.1 k &5.77 & 10.90 & 1.8 GB & 9.2 s/batch & 483.1 k \\
    w/ TCN & 4.58 &8.93 & 1.7 GB & 5.0 s/batch & 376.0 k & 4.89 & 10.75 & 1.8 GB & 5.3 s/batch & 376.0 k  \\
    w/ Time Feat. & 3.76 & 7.93 & 1.5 GB & 1.1 s/batch & 219.8 k & 4.50 & 9.90 & 1.6 GB & 1.2 s/batch & 219.8 k  \\
    Input 36 Steps & 3.73 &7.74 & 1.5 GB & 1.1 s/batch & 219.4 k & 4.52 &9.87 & 1.6 GB & 1.2 s/batch & 219.4 k  \\
    Input 48 Steps & 3.74 & 7.89 & 1.5 GB & 1.2 s/batch & 220.2 k & 4.55 &9.95 & 1.6 GB & 1.4 s/batch & 220.2 k   \\
    Input 72 Steps & 3.77  & 7.97 & 1.6 GB &  1.4 s/batch & 221.7 k & 4.53 &9.88 & 1.7 GB & 1.4 s/batch & 221.7 k   \\
    \bottomrule
    \end{tabular}}
\end{small}
\end{table}
}

{
\subsubsection{Online Prediction Performances and Updating Strategy}
It is important to consider a real-world scenario in which data is accumulated over time. As a result, the model will likely require regular updates and maintenance in order to maintain its accuracy. 
We are interested in how the model will be managed and updated with new data.
Accordingly, we utilize a single pretrained model derived from a limited number of observed regions and deploy it to online prediction tasks involving an increasing number of novel regions. 
Fig. \ref{fig:online_application} shows the online prediction results in two cities.
It is noted that this figure is different from the result in Fig. \ref{fig:different_n_u} as only a single model is applied. Our model shows better robustness than the baseline model IGNNK in the online application stage. 

Since the model is trained in an inductive manner, it can deal with increasing data scales and region numbers. The requisite accuracy can be achieved through the implementation of model fine-tuning or retraining. 
Additionally, the LLM-based encoding can be precalculated for all possible regions offline. This makes the online application efficient.

\begin{figure}[!htbp]
  \centering
  \captionsetup{skip=1pt}
  \includegraphics[width=0.75\columnwidth]{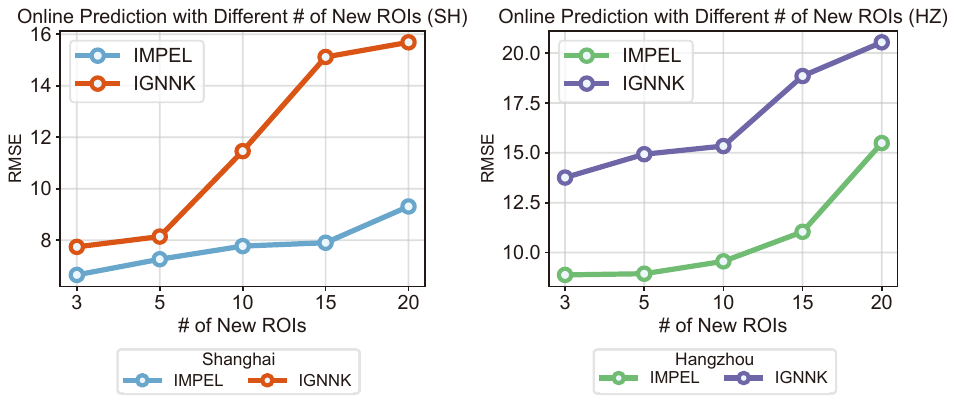}
  \caption{{Online prediction performance with different numbers of newly added regions.}}
  \label{fig:online_application}
\end{figure}

{
Regarding the online test efficiency, we evaluate the computational efficiency at both inference stage (with generated LLM-based encoding) and the LLM encoding stage. Table \ref{tab:llm-time} reports the total running time and memory consumption at two different stages. It is observed that the encoding stage requires a high computational burden due to the high complexity of LLMs. Fortunately, this encoding step (a single forward pass) only needs to be performed \textbf{once} throughout the testing or training phase and can also be performed \textbf{offline} after the location of the test points has been determined. After equipping the generated encoding into the prediction model, it can complete the testing task at low cost, preserving the capability of the online application.

\begin{table}[htbp]
\caption{{Computational efficiency of model inference and LLM-based encoding stages.}}
\label{tab:llm-time}
\centering
\begin{small}
    \renewcommand{\multirowsetup}{\centering}
    \renewcommand{\arraystretch}{0.8} 
    \setlength{\tabcolsep}{2pt}
    \resizebox{0.7\textwidth}{!}{
    \begin{tabular}{l|cc|ccc}
    \toprule
     \multirow{2}{*}{Models} & \multicolumn{2}{c|}{Testing stage} & \multicolumn{3}{c}{LLM encoding stage} \\
    \cmidrule(lr){2-6} 
    & Running Time & GPU Memory & GPU Memory & Encoding Speed & Batch Size \\
    \midrule
    \multirow{3}{*}{\textsc{Impel}} & \multirow{3}{*}{{0.17 s}} & \multirow{3}{*}{1.5 GB} & 15.3 GB & 2.48 s & 1 \\
    &  &  & 15.5 GB & 2.04 s & 4 \\
    &  &  & 15.8 GB & 2.11 s & 8  \\
    \bottomrule
    \end{tabular}}
\end{small}
\end{table}
}
}


\footnotesize
\bibliographystyle{elsarticle-harv}
\bibliography{main}



\end{document}